\documentclass[10pt,twocolumn,letterpaper]{article}

\usepackage{cvpr}
\usepackage{times}
\usepackage{epsfig}
\usepackage{graphicx}
\usepackage{amsmath}
\usepackage{amssymb}

\DeclareMathOperator*{\argmax}{argmax}

\usepackage[pagebackref=true,breaklinks=true,letterpaper=true,colorlinks,bookmarks=false]{hyperref}

\cvprfinalcopy 

\ifcvprfinal\fi
\begin{document}

\title{Joint Facade Registration and Segmentation for Urban Localization}

\author{Antoine Fond\\
Inria Grand Est \\
{\tt\small antoine.fond@inria.fr}
\and
Marie-Odile Berger\\
Inria Grand Est \\
{\tt\small marie-odile.berger@inria.fr}
\and
Gilles Simon\\
Universit\'e de Lorraine \\
{\tt\small gilles.simon@loria.fr}
}

\maketitle

\begin{abstract}
This paper presents an efficient approach for solving jointly facade registration and semantic segmentation. Progress in facade detection and recognition enable good initialization for the registration of a reference facade to a newly acquired target image. We propose here to rely on semantic segmentation to improve the accuracy of that initial registration. Simultaneously we aim to improve the quality of the semantic segmentation through the registration. These two problems are jointly solved in a Expectation-Maximization framework. We especially introduce a bayesian model that use prior semantic segmentation as well as geometric structure of the facade reference modeled by $L_p$ Gaussian Mixtures. We show the advantages of our method in term of robustness to clutter and change of illumination on urban images from various database.
\end{abstract}

\section{Introduction}

Urban localization plays  a major role in  many applications including
navigation  aid  \cite{Krolewski2011},  labeling  of local touristic
landmarks     \cite{Chen2011,Xu2002},    and     robot    localization
\cite{Wendel2011}.

Tracking solutions like  GPS can satisfy the demand to  some degree in
outdoor  environments,   but  are  prone  to   inaccuracy  in  several
situations, \eg in areas where the  street is flanked by buildings on
both sides.  Furthermore, the outdoor  accuracy of mobile phone GPS is
only 12.5 meters \cite{Zandbergen2011}  and drift-free inertial system
solutions are economically not feasible.

Image-based solutions are prone to  be more robust and accurate. These
solutions  generally rely  on  a two-step  process.   First, a  coarse
estimate  of   the  camera   pose  is  obtained \eg from   a  GPS
\cite{Arth2015,Chu2014},   user   information   \cite{Reitmayr06}   or
content-based image  retrieval \cite{Fond2017,Schindler2007}.  Second,
a 2D projection of a 3D model of the buildings visible in the image is
computed based on the coarse pose and  the pose is refined so that the
difference between the 2D projection and the real image is reduced.

This paper  is concerned with  the second  step of this  process.  The
method described in \cite{Fond2017} is  used to automatically detect a
perspective-distorted facade in a  view, recognize this facade between
a  collection  of  pre-acquired fronto-parallel  images  of  reference
facades, and compute  a coarse estimate of the pose  relatively to the
detected  facade.   From  this  estimate, we  perform  accurate  3D-2D
model-image registration  between the reference facade  and the target
image. When the  reference facade is part of  a Geographic Information
System  (GIS) our method  can  propose an  accurate  camera pose  for
geo-localization in the sense of \cite{Arth2015,Chu2016}.

Previous approaches for 3D-2D registration  based on a coarse estimate
of   the   pose   have    relied   on   invariant   feature   matching
\cite{Lowe2004,Simon2011},     correlation-based    point     matching
\cite{Simon2011},  edge  tracking  \cite{Reitmayr06} and  image  patch
tracking  \cite{Benhimane2004}.  However,  false  matches or  tracking
errors  often   occur  due   to  the  presence  of repetitive  structures
(\eg windows on  a facade),  cluttering objects,  and/or changes  in
appearance between the reference texture  and the target image, due to
different weather  conditions, time of day,  camera response function,
etc.  As a result, the number of matches is sometimes too small or the
rate of outliers too high to make any robust pose estimation algorithm
work (see \eg Fig.  \ref{fig:sift}).

The authors  of \cite{Kobyshev2014}  tried to  tackle these  issues by
including  semantic information  in  the matching  process.  In  their
method, a semantic histogram is  built to capture the semantic context
around  each detected  feature point.   These histograms  are used  to
learn which  features are  likely to match  correctly and  leaving the
other features out. This method is of limited interest in our case, as
the  considered  semantic   labels  (``road'',  ``sky'',  ``objects'',
``vegetation'', etc.)   are generally  not part of  a facade.   It may
help to discard false matches due  to cluttering objects in front of a
facade,  but  the  other  issues (repetitive  structures,  changes  in
appearance) remain.

Still, relying on a semantic segmentation to register a facade texture
in a target image has several  advantages. First, there is no need for
a complex  similarity metric as semantic  segmentation already manages
appearance   and   viewpoint   changes    between   the   two   images
\cite{Castaldo2015}.  Second,  the registration focuses  on meaningful
components on both images  reducing possible local minima. Eventually,
compared  to global  feature-based  methods it  can  benefit from  an
initial detection.  On the other hand  semantic segmentation can
still  be noisy  with  many misclassified  pixels.   Actually the  two
problems  are  linked.  Given   a  better  semantic  segmentation  the
registration can be  more accurate but when the  registration is close
to the optimal solution, it  can help to disambiguate between semantic
classes.   For example,  if a  door  is misclassified  as window,  the
labels   layout  of   the   registered  reference   can  correct   the
segmentation.  Based  on  an Expectation-Maximization  framework,  our
method aims  to solve these two  problems jointly to benefit  from one
another.

\section{Previous work}

\subsection{Semantic Image Segmentation}

Convolutional   Neural    Networks   (CNNs)    \cite{Bansal2017}   and
particularly  Fully  Convolutional   Networks  (FCNs)  \cite{BadrinarayananH15,Long2015,Noh2015}  have   proven     efficient   for
pixel-wise semantic  segmentation.  FCNs are based  on encoder-decoder
architectures  that  do  not  contain any  fully-connected  layers  or
multi-layer perceptron  (MLP) usually  found at the  end of  the CNNs.
For instance,  in the SegNet network  \cite{BadrinarayananH15} that is
used in our method, the  encoder network is topologically identical to
the 13  convolutional layers of the VGG16  network \cite{Simonyan14}.
The role of  the decoder network is to map  the low resolution encoder
feature  maps to  full input  resolution feature  maps for  pixel-wise
classification.  The pooling indices  computed in the max-pooling step
of  the   corresponding  encoder   are  used  to   perform  non-linear
upsampling. FCNs are  fast and well-suited  to online applications.   However, the
segment boundaries  can still be  noisy (see \eg Fig.   \ref{fig:rescalesem})
 and they still can missclassify very visually ambiguous classes like doors and windows.

 To improve the results of generic semantic segmentation approaches on facades,
  some authors have proposed specific methods that exploits  the facade structure.
Among the bottom-up approaches, Gadde \etal \cite{gadde17} iteratively refine the segmentation using auto-context descriptors
that enforce the rectangular-shape of the segment as well as their vertical and horizontal repetitions.
A facade segmentation made of  strictly rectangular structures can be
obtained by using  the method presented in  \cite{Yang2012}. The clutter-free low-rank texture of the facade from TILT \cite{Zhang2012}  is
initially segmented then partitioned into  multiple blocks of rank-one matrix by a heuristic split and merge approach.
There are also top-down methods that parse facades using shape grammar. In \cite{TeboulKSKP13} Teboul \etal use reinforcement learning techniques
on a Markov Decision Process to find the optimal parsing tree of the facade.
All these facade-specific approaches require the perfect boundaries of the facade and are computationally expensive. For these reasons there are poorly suited to support registration.

\subsection{Semantic-based Model-Image Registration}

Model-image  registration in urban environment  has  been  performed  based   on  semantic
segmentation in at least two previous works. In \cite{Arth2015}, an approximate
pose  provided by  a GPS  is refined  by fitting  vertical corners  of
buildings (obtained from a 2D  city map) with vertical lines extracted
from the image  (edges pointing toward the  vertical vanishing point).
However, as  the images are very  cluttered in practice, this  task is
very  challenging and  often leads  to inaccurate  registration.  This
problem is  tackled by  generating several translation  hypotheses for
each possible pair of correspondences between the building corners and
the  vertical lines  extracted from  the image.   A simple  pixel-wise
segmentation  of the  input  image is  then used  to  select the  best
translation among the hypotheses. A SVM classifier is applied to each
image patch  of a given  size to assign  a class label  ({\em facade},
{\em  sky}, {\em  roof}, {\em  vegetation}  and {\em  ground}) to  the
center location  of the patch.  The  refined pose is then  obtained by
maximizing a log-likelihood which is high when the pixels lying on the
projection of the facades have a high probability to be on a facade in
the image, and the pixels lying outside have a high probability to not
be on a  facade.

Though  this  method  is interesting,  the  accuracy of  the
registration relies on the pixel-wise segmentation, which is noisy and
do not separate adjacent facades. Moreover, structural elements on the
facades  (windows, doors,  etc.) are  not detected  by the  classifier
(they are  simply classified as  {\em facade}), though  these elements
would be useful to get a more accurate registration.

Chu \etal \cite{Chu2016} exploit this structural information to better estimate the camera location as well as
some geometric parameters of the building's model (height of each floor, vertical positions of windows and doors, etc.).
As in \cite{Arth2015}, the method assumes the camera pose to be initialized by GPS and requires geo-referenced footprint of buildings as a base for creating the 3D models.
The  problem is formulated as inference in a Markov random field, which encourages the projection of the 3D model
to   match    the   image   edges,   semantics    (based   on   SegNet \cite{BadrinarayananH15}) and location of  doors and windows (based on
Edgeboxes \cite{Zitnick2014}  and AlexNet \cite{Krizhevsky2012}) and to differ from the background in all GoogleStreetView images around the building.
Nevertheless the complexity of the inference that use a discretized parameters search space and multiple views are disadvantages for real time application to urban localization.

In  both   of  these   works  \cite{Arth2015,Chu2016},   the  semantic
segmentation is  performed once and for  all, and serves as  a basis for
the 3D-2D  registration.  However, as  we argued in  the introduction,
segmentation and  registration are  linked, and  conjointly performing
these two tasks may help improving the accuracy of both. We first propose a way to initialize both problems.
Then we introduce the bayesian model that joint them together. The details of the inference through Expectation-Maximization are described before discussing results on various databases.

\section{Initialization}

Initialization of the Expectation-Maximization procedure is based on four steps:
(i) the camera intrinsic parameters are computed from the image content
(ii) the image is rectified so that the facades of the buildings appear as if they where fronto-parallel to the camera (several rectified images can be obtained),
(iii) facades in the rectified images are detected (approximate bounding boxes are obtained) and recognized,
(iv) semantic segmentation and registration are initialized from the bounding boxes of the recognized facades.
In the following, this initialization step will be referred to as $t=t_0$. We now detail each of its subtasks.

\subsection{Autocalibration and plane rectification}

Steps (i) and  (ii) of the initialization process  are performed using
the method described in \cite{simon16}. Horizontal vanishing points of
the  image  are  detected  by  exploiting  accumulations  of  oriented
segments around the  horizon line.  The principal point  is assumed to be at
the  center of  the image  and  the focal  length is  computed from  a
detected  pair  of orthogonal  vanishing  points.   Finally, for  each
detected vanishing point a homography is computed, that transforms all
vertical planes in the direction  of the vanishing point to a
fronto-parallel view of the planes.

\subsection{Facade detection and recognition}

Facades are detected and recognized  in the rectified images using the
method presented in \cite{Fond2017}.  This method relies on image cues
that  measure typical  facade  characteristics such  as shape,  color,
contours,  structure  (windows  and  balconies are  detected  using  a
modified  version of  SegNet  \cite{BadrinarayananH15}), symmetry  and
semantic contrast.  These cues are  combined to generate a  few facade
candidates fast.   The candidates are then  classified into ``facade''
and  ``non facade''  through a  neural network  using SPP  descriptors
\cite{He2014}.   The remaining  facades  are matched  with the  facade
database  using a  semantic metric  learned through  a siamese  neural
network taking the SPP descriptors as inputs.

\subsection{Registration and segmentation initialization}

In  this method  we  aim  to jointly  solve  the  registration of  the
recognized  reference to the detected  facade  in the target image  and  the
segmentation of the latter into semantic  parts. As the image has been
previously  rectified  using  calibrated camera  intrinsics  the  only
remaining parameters to  register the reference image  onto the target
image are one scale parameter $s$  (the aspect ratio is preserved) and
two  translational  parameters  $\left(   t_x,  t_y  \right)$.  Facade
recognition  enables to  select  the correct  facade  reference to  be
registered in  a larger  facades database.  Moreover thanks  to facade
detection we can  estimate a first initialization  of the registration
parameters by solving the least-square problem that maps the four transformed corners
 of the reference to the four corners of the detection.

\begin{figure}[h!]
\includegraphics[width=0.5\linewidth]{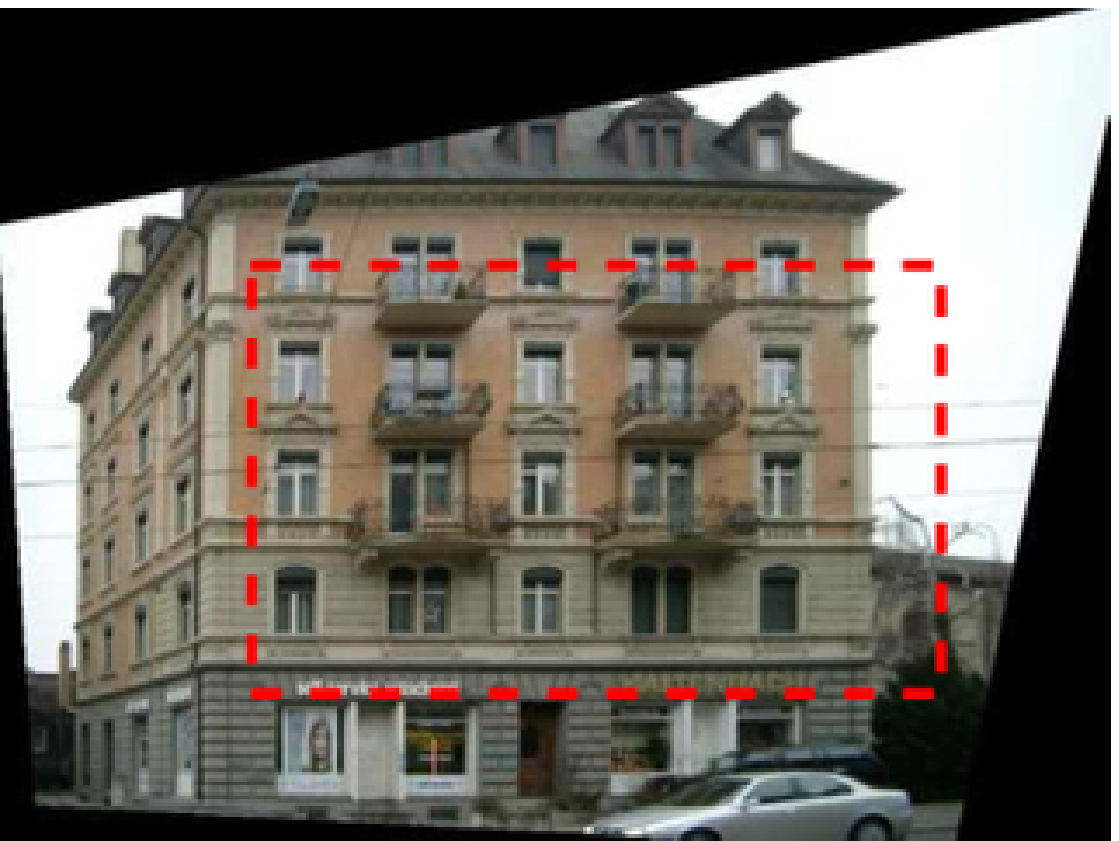}\includegraphics[width=0.5\linewidth]{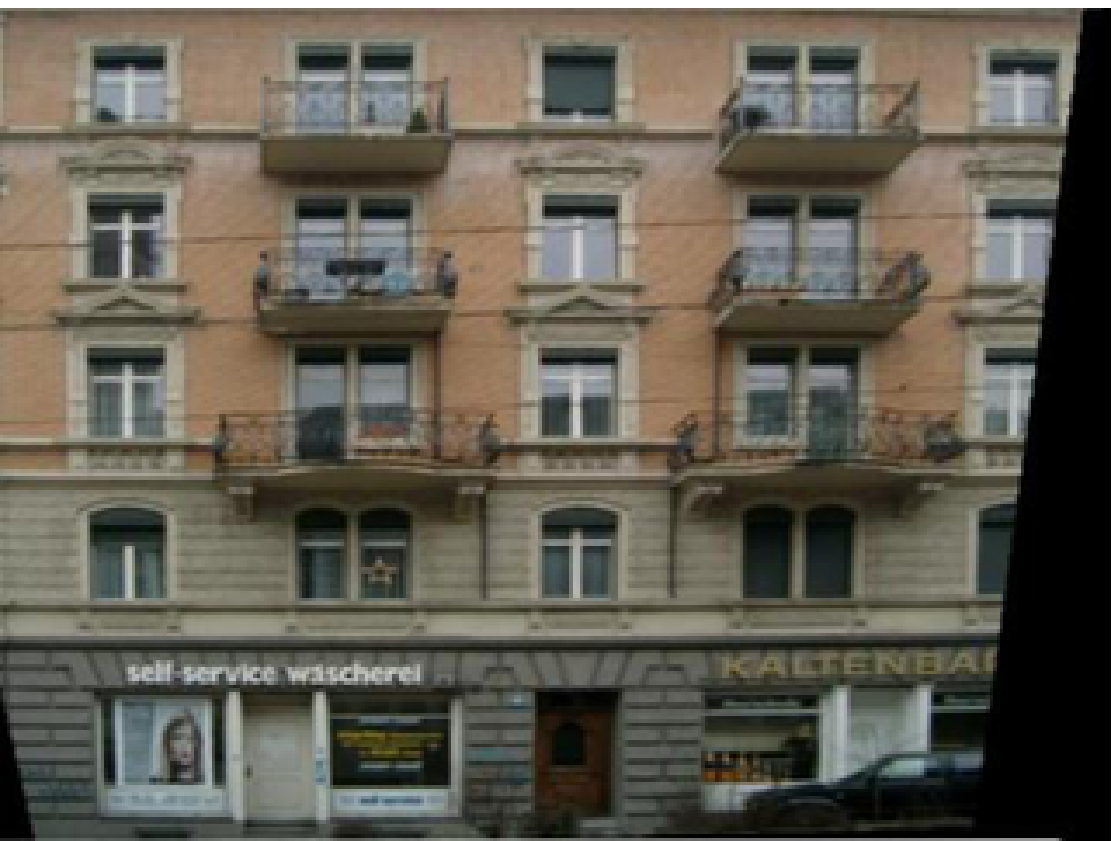}
\caption{The initial registered boundaries of the reference $I_{ref}$ (right) overlay the target image $I$ (left).}
\label{fig:registration}
\end{figure}

\begin{figure}[h!]
\includegraphics[width=0.498\linewidth]{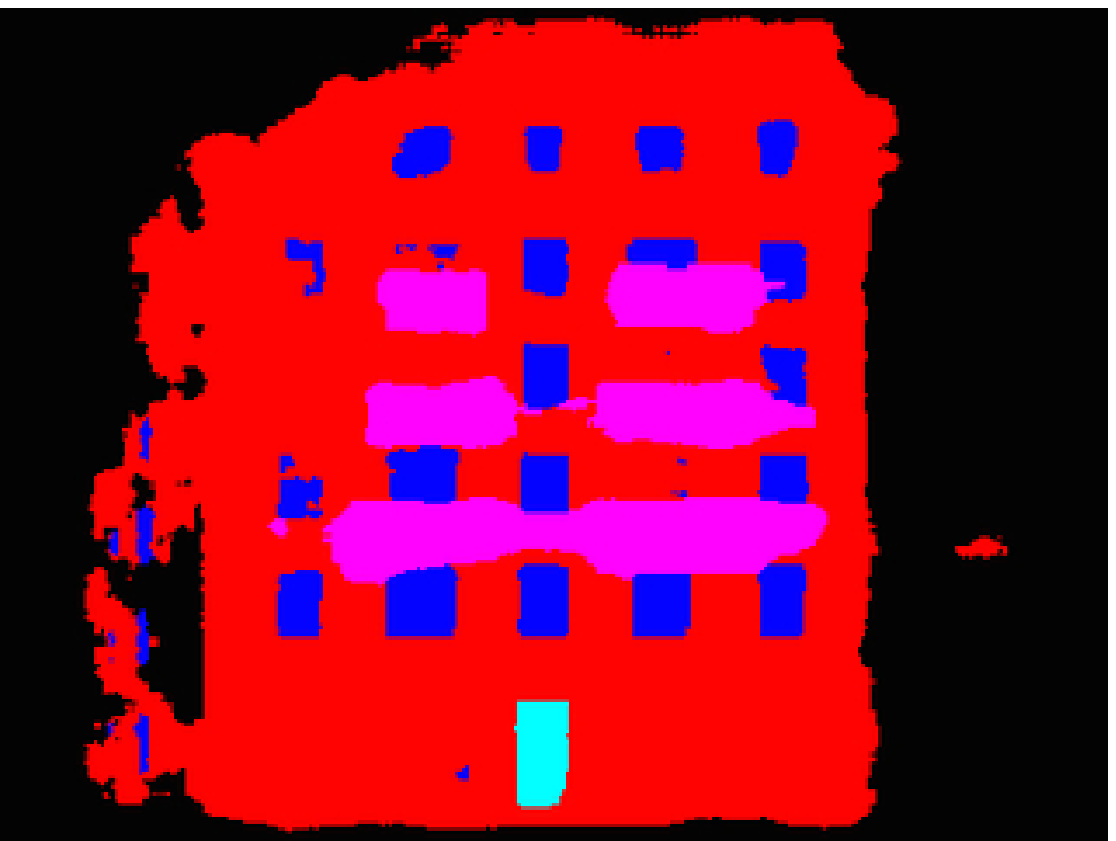}\includegraphics[width=0.5\linewidth]{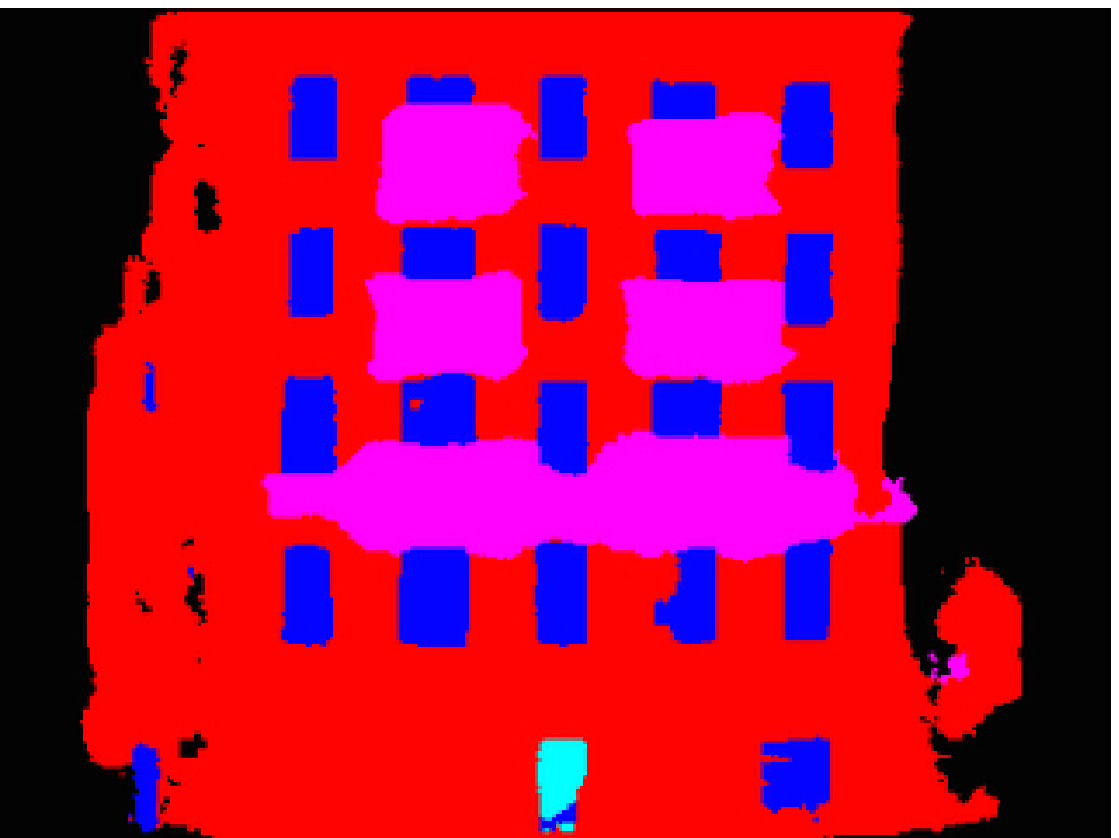}
\caption{Semantic segmentation initialization of the target image $I$ without rescaling (left) and with rescaling (right).}
\label{fig:rescalesem}
\end{figure}

As the facade detection step relies on semantic segmentation, it also provides a first initialization of the latter. However its SegNet \cite{BadrinarayananH15} inference is sensitive to scale (Fig. \ref{fig:rescalesem}). To improve this initial segmentation, we zoom in the image region corresponding to the transformed boundaries of the reference and we perform another inference. The transformation uses the estimated scale $s$ augmented by a constant to avoid the region of interest to be to much cropped.

\section{Joint registration and semantic segmentation}

\subsection{Bayesian model}

\begin{figure}[h!]
\centerline{\includegraphics[width=0.7\linewidth,height=3.5cm]{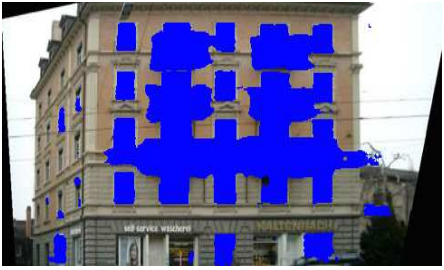}}
\caption{Data points $X$ from the target image $I$. Only the points from pixels which are likely ($P(l_j|i,I) \geq 0.01$) to be a characteristic facade architecture components are considered.}
\label{fig:data}
\end{figure}

We wish to register the recognized image reference $I_{ref}$ onto the target image $I$ in which the facade has been detected through the transformation $T$ and simultaneously improve the quality of the semantic segmentation. We denote $L = \left\{ l_j \right \}_{1 \leq j \leq K}$ the different labels from the semantic segmentation that are characteristic of a facade architecture such as "window", "door" and "balcony". The other labels from \cite{Fond2017} (i.e. "facade","sky","road","background") are not considered.  Both target and reference images are considered as sets of 2D labeled points. Let $X = \left\{ X_i \right \}_{1 \leq i \leq N}$ be a set of $N$ data points $X_i = \left( x_i, y_i \right)$ from the target image $I$. These points are the coordinates of the pixels $i$ from the target image $I$ that have a fair probability of being one of the labels $P(l_j|i,I) \geq 0.01$ (Fig. \ref{fig:data}). This probability $P(l_j|i,I)$ is the score of the last layer of the CNN for semantic segmentation. The set of points $X_{ref}$ from the reference image $I_{ref}$ is modeled by a mixture of $L_p$ gaussian distributions $\mathcal{N}_p$ (Eq. \ref{eq:gmm})  for each label $l_j$: $\left( \pi_{k_j} \,, \mu_{k_j} \,, \Sigma_{k_j} \right)_{1 \leq k_j \leq m_j}$. Those distributions are well suited for facade architectural components as the $L_p$ norm $\left\|  M \right\|_{p,\Sigma}^p =  \frac{m_x^p}{\Sigma_{xx}} + \frac{m_y^p}{\Sigma_{yy}}$ unit ball is roughly rectangular with a high value of $p$. The goal is to estimate the geometric transformation $T(\Theta)$ of parameters $\Theta = \left( t_x, t_y, s \right)$ that registers these $L_p$ gaussians to the set of observed data points $X$ from the target image $I$.  In addition, the assignment of a data point $X_i$ to a transformed $L_p$ gaussian as well as the prior segmentation probability $P(l_j|,i,I)$ can be seen as a posterior segmentation. Assuming that the observed data $X$ are independent and taking the logarithm, the \textit{a posteriori} distribution can be maximized to find $\Theta$ :

\begin{equation}
\label{eq:map}
\begin{split}
\Theta^\star &= \argmax_{\Theta} \ln P(X|\Theta,I)P(\Theta) \\
&= \argmax_{\Theta} \sum_{i=1}^N \ln P(X_i|\Theta,I) + \ln P(\Theta)
\end{split}
\end{equation}

\begin{figure}[h!]
\includegraphics[width=0.5\linewidth]{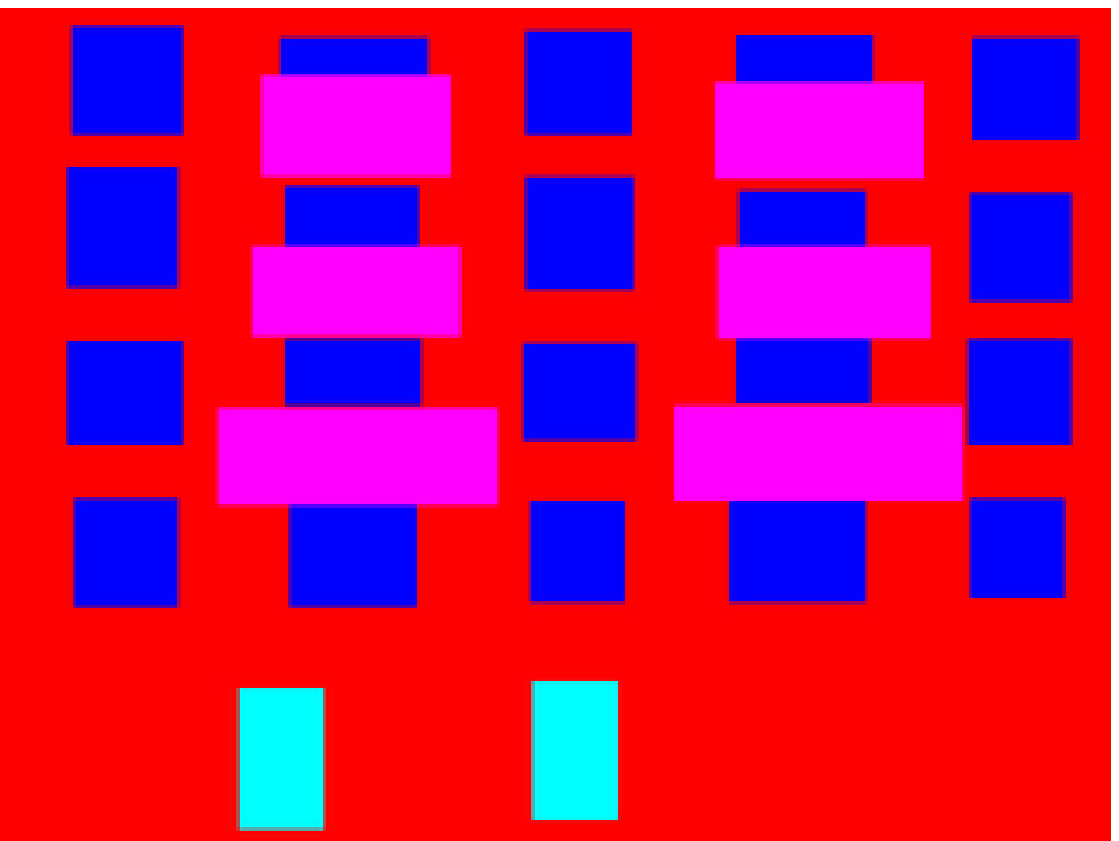}\includegraphics[width=0.5\linewidth]{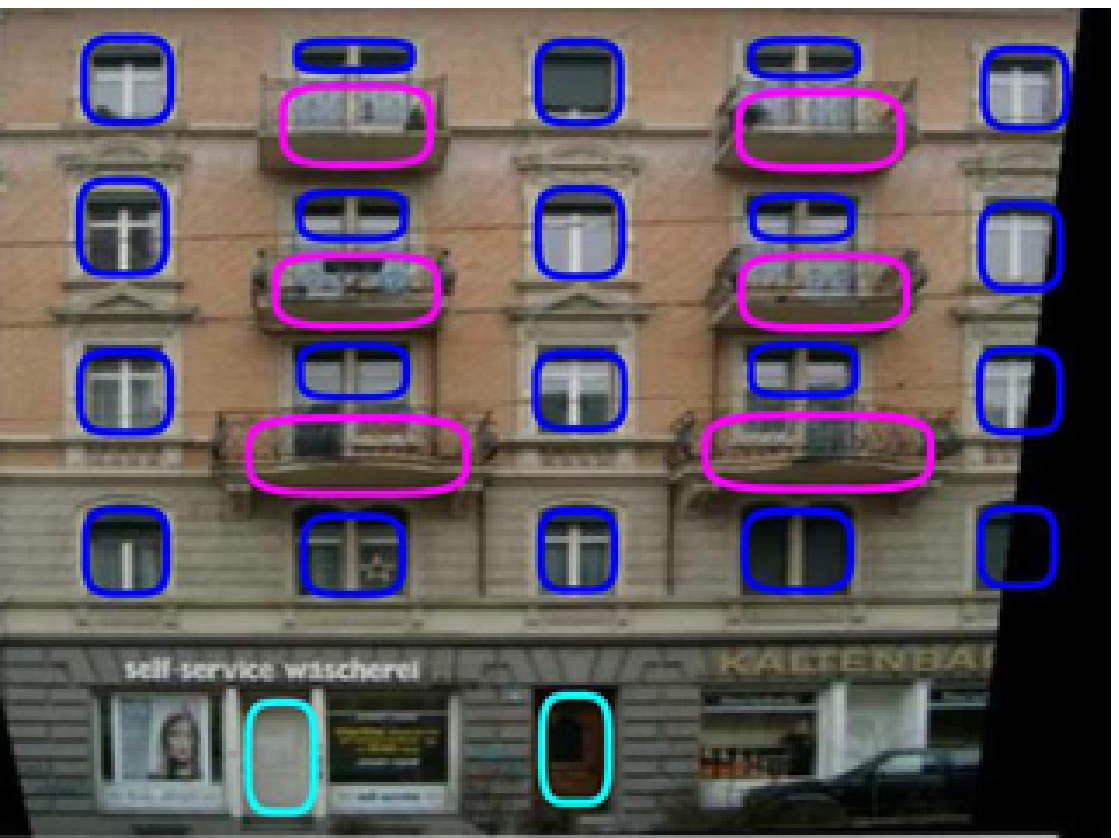}
\caption{Ground-truth of the semantic segmentation from the reference image $I_{ref}$ (left) and the $L_p$ gaussians mixtures to model it (right).}
\label{fig:gaussians}
\end{figure}

Using the law of total probability, we can introduce the probability $P(X_i|l_j,\Theta,I)$ which is modeled by a mixture of transformed $L_p$ gaussians (Eq. \ref{eq:gmm}), and $P(l_j|i,\Theta,I)$ which can be seen as a segmentation prior probability:
\begin{equation}
\label{eq:labels}
\begin{split}
P(X_i|\Theta,I) &= \sum_{j=1}^K P(X_i|l_j,\Theta,I) P(l_j|i,\Theta,I) \\
&+ P(X_i|o,\Theta,I) P(o|i,\Theta,I)
\end{split}
\end{equation}
$\alpha = P(o|i,\Theta,I)$ is the outliers rate and we choose a spatial uniform distribution to model outliers predictions $P(X_i|o,I) = \frac{1}{H W}$ with $H,W$ the dimensions of the target image. In practice, the outliers rate is initialized to $\alpha=0.25\left(1-\frac{s^2 h w}{HW}\right)$ with $h,w$ the dimensions of the reference. Moreover thanks to the scale reestimation and the invariance of CNN to small translations, the semantic segmentation inference is pretty stable. Thus we can assume that $P(l_j|i,\Theta,I) = P(l_j|i,\Theta^{(t_0)},I)$.

\begin{equation}
\label{eq:gmm}
\begin{split}
P(X_i|l_j,\Theta,I) &= \sum_{k_j=1}^{m_j} \pi_{k_j} \mathcal{N}_p \left( X_i | T \mu_{k_j}, s^p \Sigma_{k_j} \right) \\
&= \sum_{k_j=1}^{m_j} \pi_{k_j} \frac{\exp \left( - \left\| X_i - T \mu_{k_j} \right\|_{p,s^{p} \Sigma_{k_j}}^p  \right) }{4/p^2 \Gamma (1/p)^2 | s^{p} \Sigma_{k_j} | }
\end{split}
\end{equation}
To properly model the rectangular shape of facade components and keep the computation tractable we choose $p=4$.
The number of $L_p$ gaussians and their parameters are set from the image reference $I_{ref}$ (Fig. \ref{fig:gaussians}). First, we suppose that the ground-truth semantic segmentation of the image reference of the detected facade is already available. Then, for each label $l_j$, we extract the connected components and a $L_p$ gaussian is fitted in each of them. As the image is rectified and the shape of the connected component is typically rectangular, the axis of the $L_p$ gaussians are aligned with the image axis. The center of the $L_p$ gaussian $\mu_{k_j}$ is initialized to the mean of the pixels coordinates of the connected component and the covariance $\Sigma_{k_j} = $ diag $\left( \sigma_x^{p/2}, \sigma_y^{p/2} \right)$ is initialized from their vertical and horizontal variance (respectively $\sigma_x$ and $\sigma_y$). They are then refined by minimizing the error between the connected component and the true $L_p$ gaussian form using Gauss-Newton. The mixture priors $\left( \pi_{k_j} \right)_{1 \leq j \leq K \,,1 \leq k_j \leq m_j}$ are initialized such as $\pi_{k_j}$ is the ratio of the number of points $X_{ref}$ from the connected component $k_j$ over the total number of points $X_{ref}$ from the image reference $I_{ref}$. Then they are normalized $\sum_{j,k_j} \pi_{k_j} = 1$.

To be more robust to clutter we let the mixture weights free to vary during the inference but, as a tradeoff, we assume a prior distribution over them.
We can actually add the mixture weights to the parameters $\Theta = \left( \{\pi_{k_j}\}_{1 \leq j \leq K \,,1 \leq k_j \leq m_j}, \alpha, t_x, t_y, s \right)$ without changing Eq. \ref{eq:map}. We don't assume any prior for the transformation parameters $\left( t_x, t_y, s \right)$ but we choose a Dirichlet distribution as a prior for the mixture weights $\pi_{k_j}$:

\begin{equation}
\label{eq:dirichlet}
 P(\Theta) = \mathcal{D}ir \left( \pi_{k_j} | \alpha_{k_j} \right)_{1 \leq j \leq K \,,1 \leq k_j \leq m_j} \propto \prod_{j,k_j} \pi_{k_j}^{\alpha_{k_j}-1}
\end{equation}
Gauvain \etal \cite{GauvainL94} show that Dirichlet distribution is a practical prior candidate for mixture distributions that enables closed-form formula to the following equations. $\left( \alpha_{k_j} \right)_{1 \leq j \leq K \,,1 \leq k_j \leq m_j}$ are set to the same values as the initialized mixture priors $\alpha_{k_j}  = \pi_{k_j}^{(t_0)}$.

\subsection{Expectation-Maximization}
This Maximum \textit{A Posteriori} (MAP) problem can be solved in the framework of expectation-maximization. We define the latent variables $Z = \left\{ z_{i,j,k_j} \in \{0,1\}, z_{i,o} \in \{0,1\}\right\}_{1 \leq i \leq N \, , 1 \leq j \leq K \,,1 \leq k_j \leq m_j}$ such that $z_{i,j,k_j}$ assign a point $X_i$ to a $L_p$ gaussian $\left( T \mu_{k_j}, s^p \Sigma_{k_j} \right)$ from the label $l_j$ and $z_{i,o}$ assign $X_i$ to the outlier extra class $o$. The Expectation-Maximization algorithm seeks to find the solution iteratively by alternating between calculating the expected complete-data log-likelihood $Q(\Theta|\Theta^{(t)})$ with respect to $Z$ given $X$ and the current parameters $\Theta^{(t)}$ and finding the parameters $\Theta$ that maximizes this quantity :
\begin{equation}
\begin{split}
Q\left( \Theta| \Theta^{(t)} \right) &= \mathbb{E}_{Z|X,\Theta^{(t)}} \ln P(X,Z|\Theta)  \\
&= \sum_Z P(Z|X,\Theta^{(t)}) \ln P(X,Z|\Theta) \\
&= \sum_{i,j} \sum_{k_j} \beta_{i,j,k_j} \left( \ln \pi_{k_j} +  \ln P(l_j|i,I) \right) \\
&+ \sum_{i,j} \sum_{k_j} \beta_{i,j,k_j}  \ln \mathcal{N}_p \left( X_i | T \mu_{k_j}, s^p \Sigma_{k_j} \right) \\
&+ \sum_i  \gamma_i  \ln \frac{\alpha}{HW}
\end{split}
\end{equation}
with $\beta_{i,j,k_j} = \mathbb{E} \left( z_{i,j,k_j} | X, \Theta^{(t)} \right)$ and $\gamma_i = \mathbb{E} \left( z_{i,o} | X, \Theta^{(t)} \right)$ \\

Thus the Expectation-Maximization framework iterates between the two steps :

\begin{itemize}
\item \textbf{E-Step:} compute $\beta_{i,j,k_j}$ and $\gamma_i$
\item \textbf{M-Step:} $\Theta^{(t+1)} = \argmax_{\Theta} Q\left( \Theta| \Theta^{(t)} \right) + \ln P(\Theta)$
\end{itemize}

The \textbf{E-Step} can be seen as the computation of an assignment probability of each observed data point $X_i$ to a $L_p$ gaussian $\left( T \mu_{k_j}, s^p \Sigma_{k_j} \right)$ from the label $l_j$ knowing the current parameters $\Theta^{(t)} = \left( \{\pi_{k_j}\}_{1 \leq j \leq K \,,1 \leq k_j \leq m_j}^{(t)},\alpha^{(t)}, t_x^{(t)}, t_y^{(t)}, s^{(t)} \right)$. Using Bayes rule and by denoting $\lambda = \frac{\alpha}{HW}$, we can write :

  \begin{equation}
  \label{eq:estep1}
  \begin{split}
  &\beta_{i,j,k_j} = \mathbb{E} \left( z_{i,j,k_j} | X, \Theta^{(t)} \right) \\
  &= \frac{\pi_{k_j} \mathcal{N}_p \left( X_i | T \mu_{k_j}, s^p \Sigma_{k_j} \right)P(l_j|i,I)}{\sum_{j',k'} \pi_{k'_{j'}} \mathcal{N}_p \left( X_i | T \mu_{k'_{j'}}, s^p \Sigma_{k'_{j'}} \right)P(l_{j'}|i,I) + \lambda}
  \end{split}
  \end{equation}

    \begin{equation}
    \label{eq:estep2}
  \begin{split}
  &\gamma_i = \mathbb{E} \left( z_{i,o} | X, \Theta^{(t)} \right) \\
  &= \frac{\lambda}{\sum_{j',k'} \pi_{k'_{j'}} \mathcal{N}_p \left( X_i | T \mu_{k'_{j'}}, s^p \Sigma_{k'_{j'}} \right)P(l_{j'}|i,I) + \lambda}
  \end{split}
  \end{equation}

  In the \textbf{M-Step} we aim to maximize $R = Q\left( \Theta| \Theta^{(t)} \right) + \ln P(\Theta)$ knowing the assignments $\beta_{i,j,k}$ and $\gamma_i$. By replacing the expressions of the distribution from equations \ref{eq:gmm} and \ref{eq:dirichlet} and by ignoring the constant terms, $R$ can be re-written as $\tilde{R}$:

  \begin{equation}
  \label{eq:mstep}
  \begin{split}
&\tilde{R} = -\sum_{i,j,k_j} \frac{\beta_{i,j,k_j}}{2} \left( \ln |s^p \Sigma_{j,k_j}| + \left\|  X_i - T \mu_{k_j} \right\|_{p,s^p \Sigma_{j,k_j}}^p \right) \\
    &+   \sum_{i,j,k_j} \beta_{i,j,k_j} \ln \pi_{k_j} + \sum_i \gamma_i \ln \lambda + \sum_{j,k_j} \left( \alpha_{k_j} - 1 \right) \ln \pi_{k_j}
\end{split}
\end{equation}

From the partial derivatives $\frac{\partial \tilde{R}}{\partial t_x}=\frac{\partial \tilde{R}}{\partial t_y}=\frac{\partial \tilde{R}}{\partial s}=0$ we can derive a polynomial system which cannot be solved in closed-form for $p=4$. Our solving strategy is similar to the one we used in the initialization of the mixture from the reference. First we solve the polynomial system in closed-form with $p=2$ which corresponds to a gaussian mixture (cf. Appendix \ref{app:solutions}). Then we refine the result by minimizing $J = \left\| \frac{\partial \tilde{R}}{\partial t_x} \right\|^2 + \left\| \frac{\partial \tilde{R}}{\partial t_y} \right\|^2 + \left\| \frac{\partial \tilde{R}}{\partial s} \right\|^2$ for $p=4$ using gradient descent. As $J$ is polynomial both the gradient and the hessian can be computed using their polynomial expression in the Gauss-Newton algorithm. The convergence is reached after a few iterations and we can update the transformation parameters $\left( t_x^{(t+1)}, t_y^{(t+1)}, s^{(t+1)} \right)$. The update for the mixtures weights $\pi_{k_j}$ and the outliers rate $\alpha$ follows the formula from \cite{GauvainL94}:

\begin{equation}
\pi_{k_j}^{(t+1)} = \frac{\sum_i \beta_{i,j,k_j} +\alpha_{k_j}-1}{\sum_{i,k'_{j'}} \beta_{i,j,k_{j'}}+\sum_{k'_{j'}} \left( \alpha_{k'_{j'}} - 1 \right)}
\end{equation}

\begin{equation}
\alpha^{(t+1)} = \frac{\sum_i \gamma_i}{\sum_{i,k'_{j'}} \beta_{i,j,k_{j'}}+\sum_{k'_{j'}} \left( \alpha_{k'_{j'}} - 1 \right)}
\end{equation}

\section{Results}

\subsection{Implementation and efficiency}

Unlike most EM approaches, in our method the $L_p$ gaussian parameters are fixed except for the mixture prior weights. Indeed here the $L_p$ gaussians model the  semantic components of the reference facade. This compact representation of a facade enables our method to be efficient. The number of $L_p$ gaussians is in the order of the number of windows. It typically varies between 2 and 30 for european style facades. The number of data points $N$ is harder to estimate but, if we assume that the image is full of adjacent facades and the empty space between windows is as large as the window itself we can approximate $N \approx 0.25 HW$. In our testing data, this approximation is valid and the average number of data points is $\hat{N} = 31072$. Actually registration does not request the points to be sampled at each pixel. In our implementation we use a multi-resolution scheme with 2 levels. The EM algorithm is executed on a downsampled version of the set of points $X$ until convergence $\left\|  \Theta^{(t+1)} - \Theta^{(t)} \right\| \leq \epsilon$ and then executed again on the full set $X$ from the last estimated $\Theta^{(t)}$.

\begin{figure}[h!]
\centerline{\includegraphics[width=0.7\linewidth,height=3.5cm]{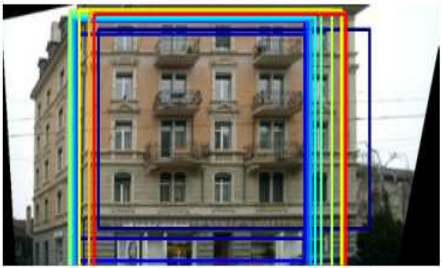}}
\caption{The registered reference boundaries of the image reference for each iteration of the algorithm are drawn in color according to the jet colormap. From dark blue for the initial iteration to red for the final one.}
\label{fig:iterations}
\end{figure}

The complexity for one iteration $t$ of the EM algorithm is $O\left(N K \max_j m_j \right)$ and parallelization is easy for the E-Step as $\beta_{i,j,k_j}$ computations are independent. This efficient complexity is also a consequence of the partial solvability of the M-Step in closed-form. The code of our implementation is mainly in Matlab but the EM is in C++. The average computation time for one iteration $t$ is $0.023$ second on an I7-3520M CPU. The number of steps for the EM to converge strongly depends on the initialization. In our testing data, only 6 iterations are needed to converge for the downsampled level and 2 more for the upper level (Fig. \ref{fig:iterations}). The computation time of the Gauss-Newton inner-iterations in the M-Step is negligible. Our optimization scheme for this step is also faster and more accurate on this problem than homotopy continuation methods. Thus the average computation time of the whole EM is 0.121 second.

To avoid the problem of the EM converging to a local maxima, we use several initializations in practice. We apply our method not only to the detected facade but also to the top-20 facade proposals \cite{Fond2017} that overlap the detected facade. The final solution is the one with the highest $R$ value.

\subsection{Validation with ground-truth semantic references}

We test our method on 3 different datasets. The first one is VarCity 3D \footnote{https://varcity.ethz.ch/3dchallenge}. It consists of 401 street-view images of buildings along the same street. Images are also semantically labelled and a 3d reconstruction of the scene is available as well as the camera parameters. 
The image viewpoints are roughly fronto-parallel and facades cover almost the whole image. As a consequence the change of scale from the reference is minor but the translation value can be high with large image parts not visible.

The second one is the 100 first buildings from Zurich Buildings Database (ZuBuD) with 5 different viewpoints per building.
Among those scenes we keep only the ones that have been correctly reconstructed by SFM \footnote{http://ccwu.me/vsfm}. 
The diversity of viewpoints in this dataset enables a wider range of scale than VarCity as well as occlusions.

The last dataset aims to show the robustness of the proposed method to change in illumination. It consists of 2 time-lapses of the same facade taken from the same viewpoint at dawn and sunrise for a total of 56 images.

For each building in all 3 databases we select the facade reference from the most fronto-parallel viewpoint where the facade is fully visible with the least occlusions possible. The reference is manually segmented into the 3 semantic labels "window", "door" and "balcony" (Fig. \ref{fig:gaussians}). The ground truth boundaries of the reference are transferred to all the images where this facade is visible using the geometric information from the SFM model.

We compare our method to both image-based and feature-based registration between the rectified target image and the reference image. In the first category we are competing against raw detection \cite{Fond2017}, $L_2$ norm minimization between images by gradient descent, Mutual Information maximization \cite{mattes2001}\cite{smriti2005}, and phase correlation \cite{reddy1996}. For the optimization methods the same initializations as for our method are chosen. For the feature-based method we extract SIFT descriptors in the rectified image with fixed orientation origin. 3 pairs of matched SIFT descriptors using Lowe's criteria \cite{Lowe2004} are used to generate transformation samples in a RANSAC framework. The comparison is done in the image itself computing the cumulative normalized histogram of the error in translation and scale. For ZuBuD and VarCity 3D the SFM models enable us to also show the error on the camera pose translation deducted from the registration (Tab. \ref{tab:3derror}).

\begin{figure}[h!]
\centerline{\includegraphics[width=0.5\linewidth]{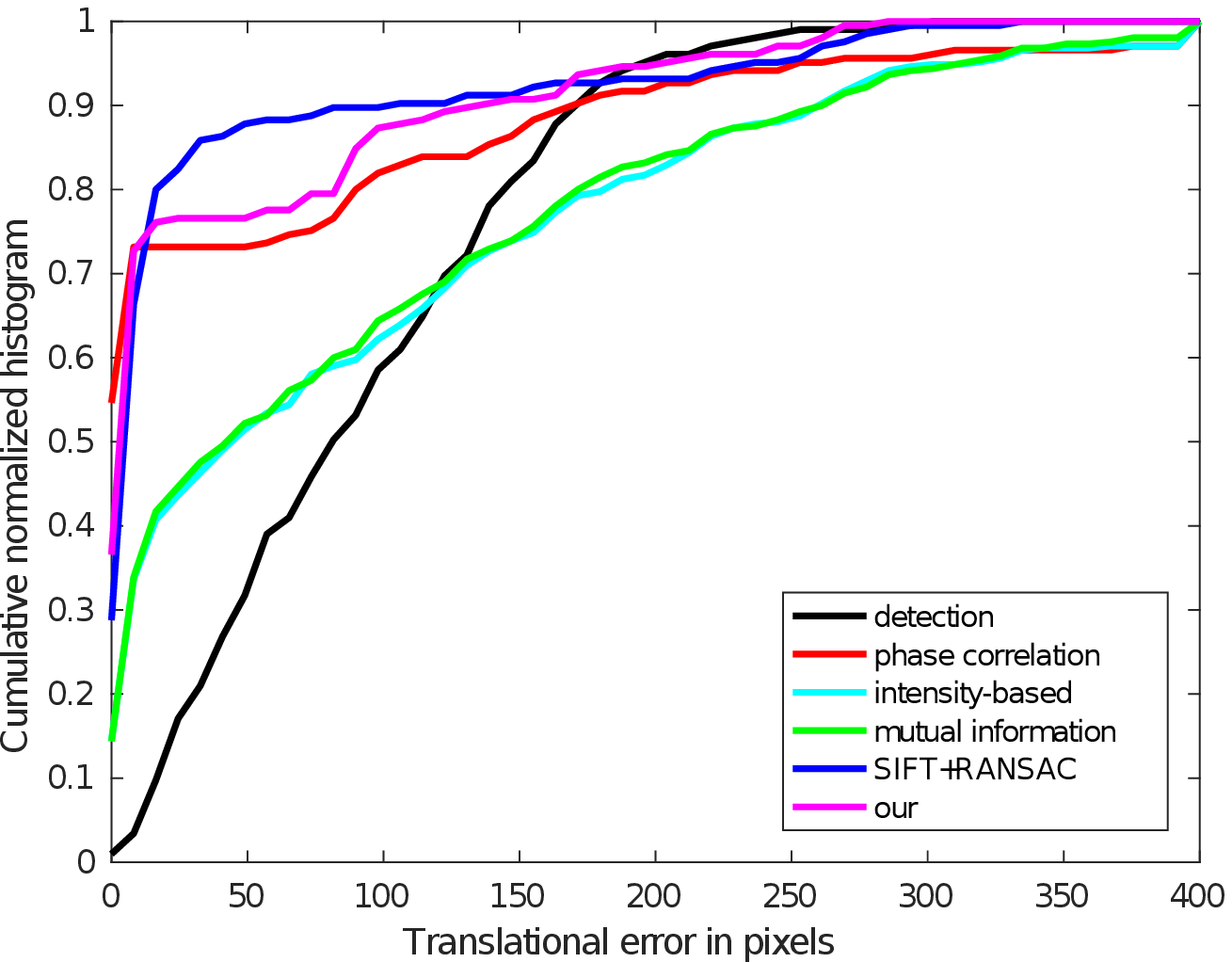}\includegraphics[width=0.49\linewidth]{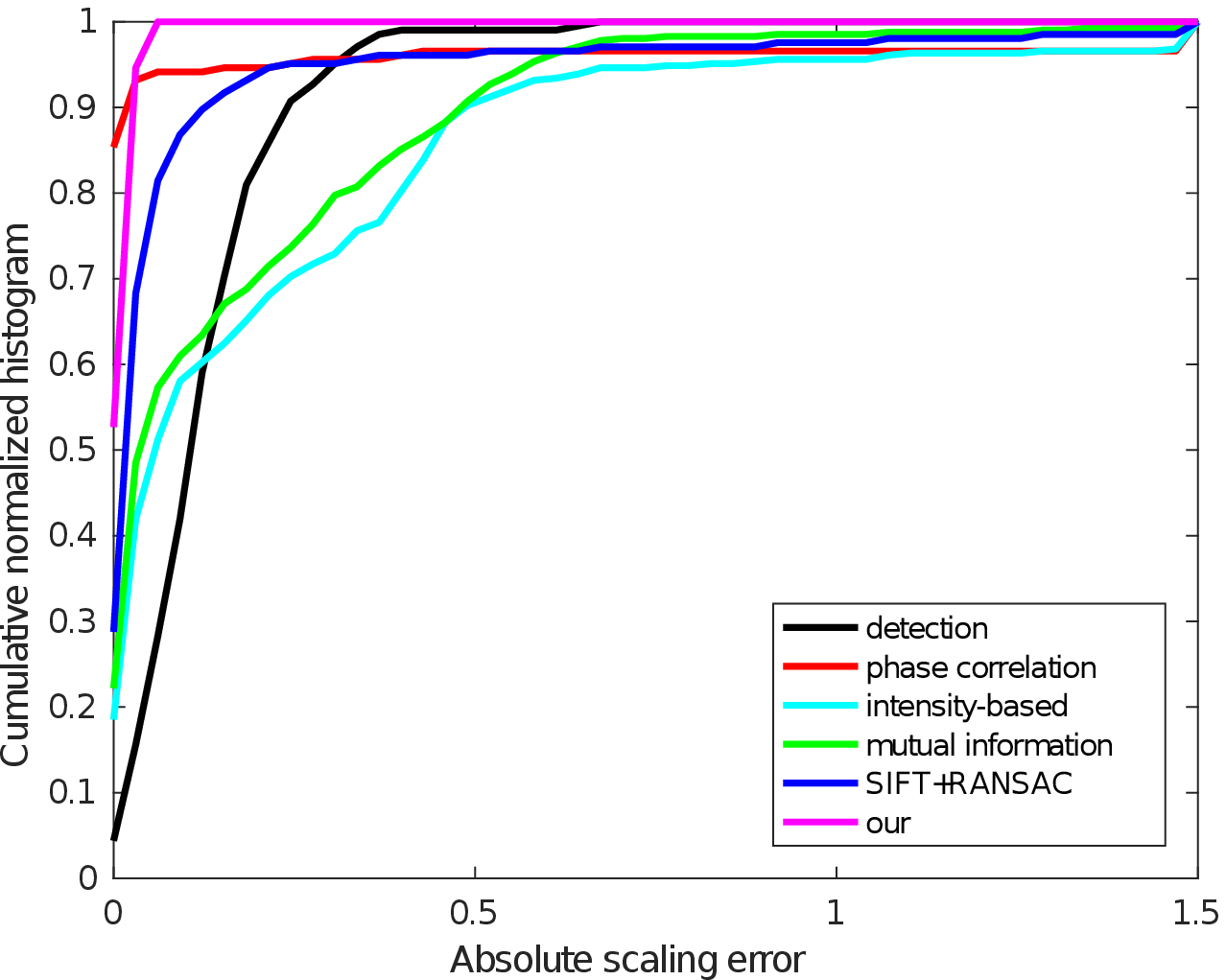}}
\caption{Registration errors in Varcity 3D}
\label{fig:varcityres}
\end{figure}

\begin{figure}[h!]
\centering
\includegraphics[width=0.99\linewidth,height=2.8cm]{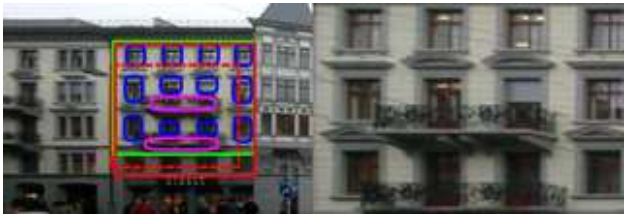}
\caption{Intensity-based approach (red) fails to estimate the registration being stuck in a local minima, whereas our method (green) succeeds. The initial (dashed line) and final (plain line) registered reference boundaries overlay the target image.}
\label{fig:esm}
\end{figure}

The good results on VarCity 3D (Fig. \ref{fig:varcityres}) show that our method can handle large translations thanks to the infinite $L_p$ gaussian support. Even when this phenomenon concurs with very repetitive patterns, the multiple initializations that exploit those repetitions and symmetries as well as the MAP regularization globally provide a correct registration. On the contrary this is a major drawbacks of methods using template-based optimization that get easily stuck in a local minima in such cases (Fig. \ref{fig:esm}). Still, sometimes in our method, the lack of discriminative architectural components like doors can cause the same shift in registration aligning the wrong floor or windows when SIFT can handle it using other features.

\begin{figure}[h!]
\centerline{\includegraphics[width=0.5\linewidth]{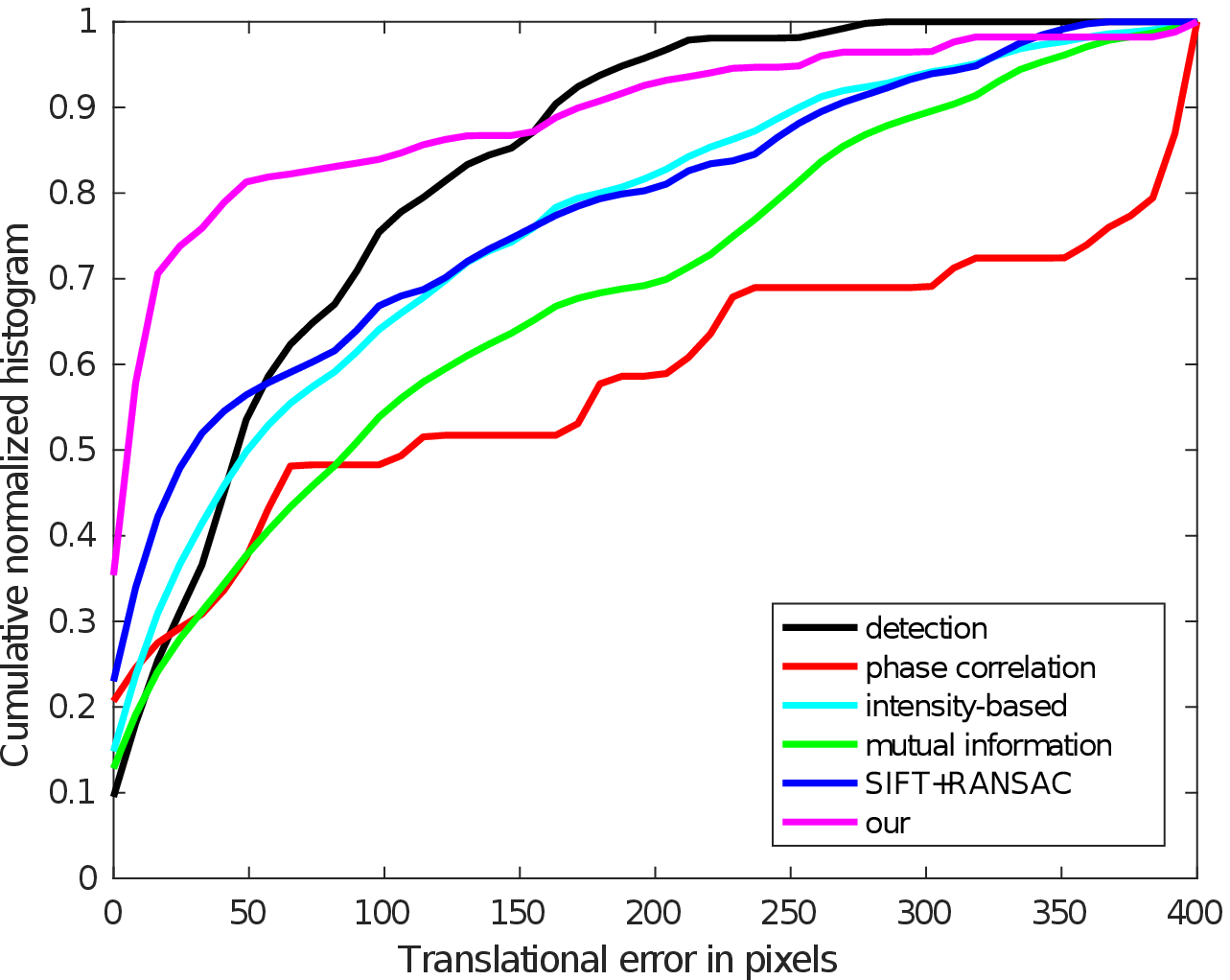}\includegraphics[width=0.5\linewidth]{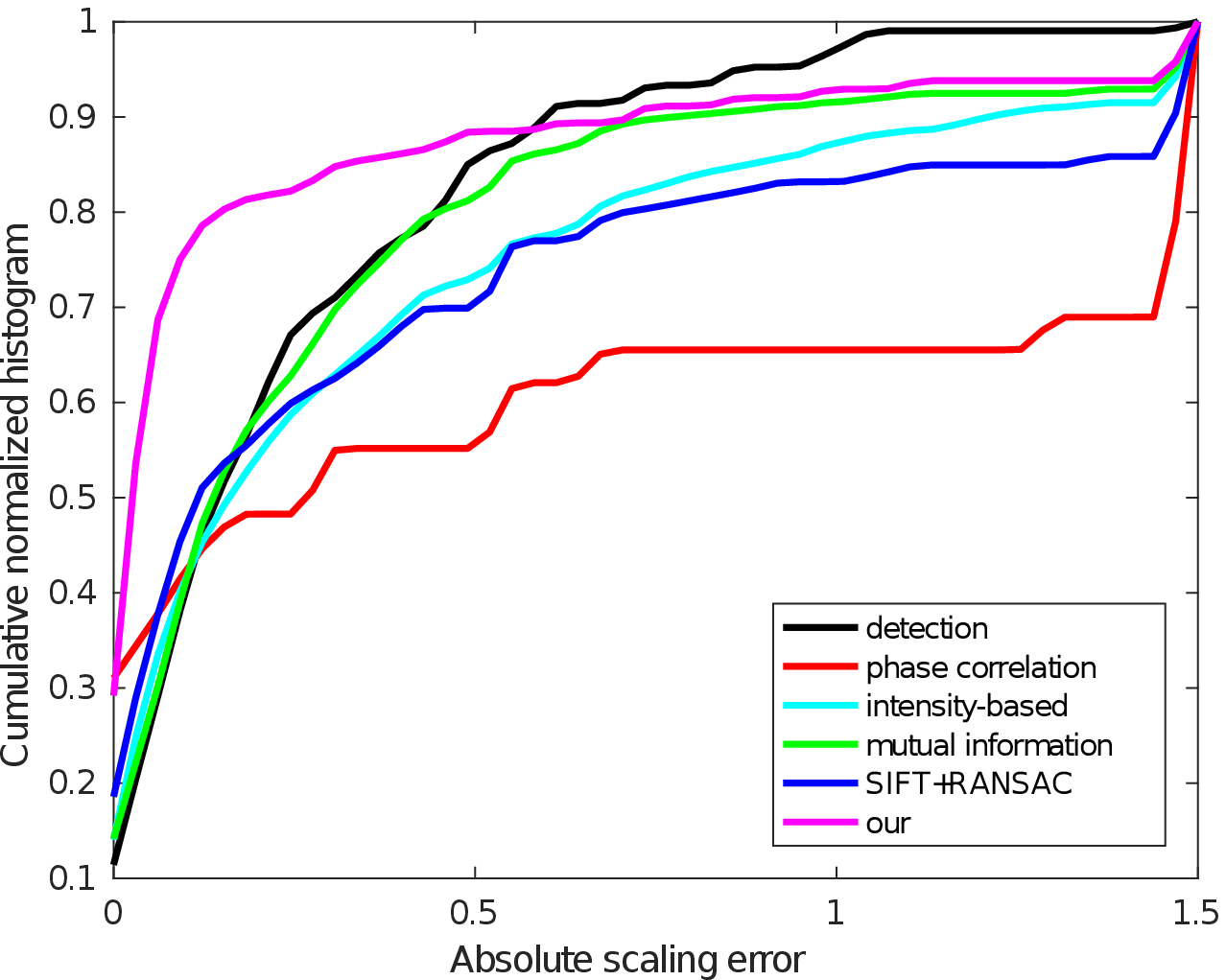}}
\caption{Registration errors  in ZuBuD}
\label{fig:zubudres}
\end{figure}

\begin{figure}[h!]
\centering
\includegraphics[width=\linewidth,height=2.8cm]{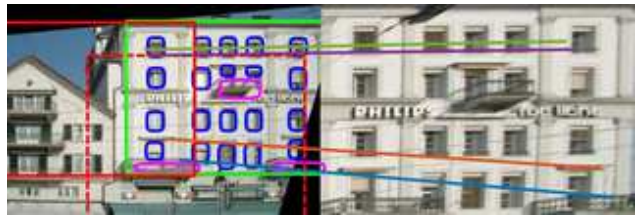}
\caption{SIFT/RANSAC approach (red) fails to estimate the registration because of the facade symmetry, whereas our method (green) succeeds. The initial (dashed line) and final (plain line) registered reference boundaries overlay the target image.}
\label{fig:sift}
\end{figure}

\begin{figure}[h!]
\includegraphics[width=0.245\linewidth,height=2.8cm]{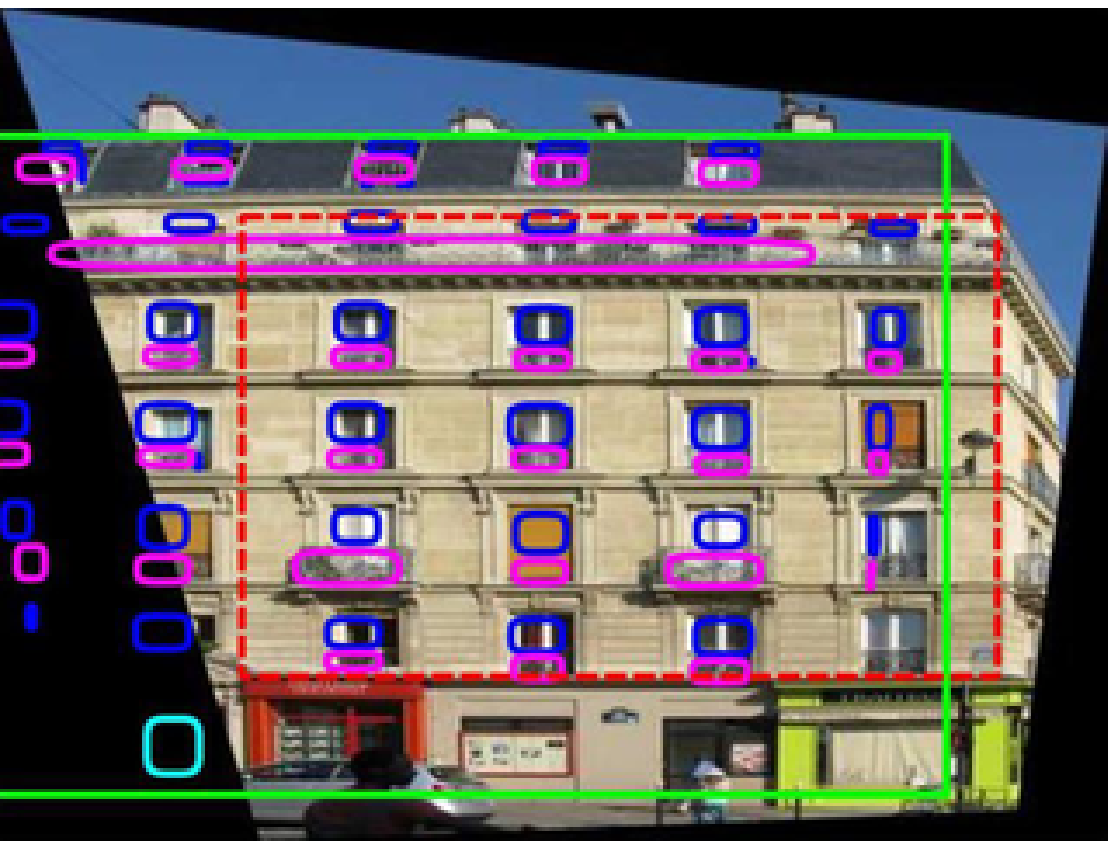}\includegraphics[width=0.245\linewidth,height=2.8cm]{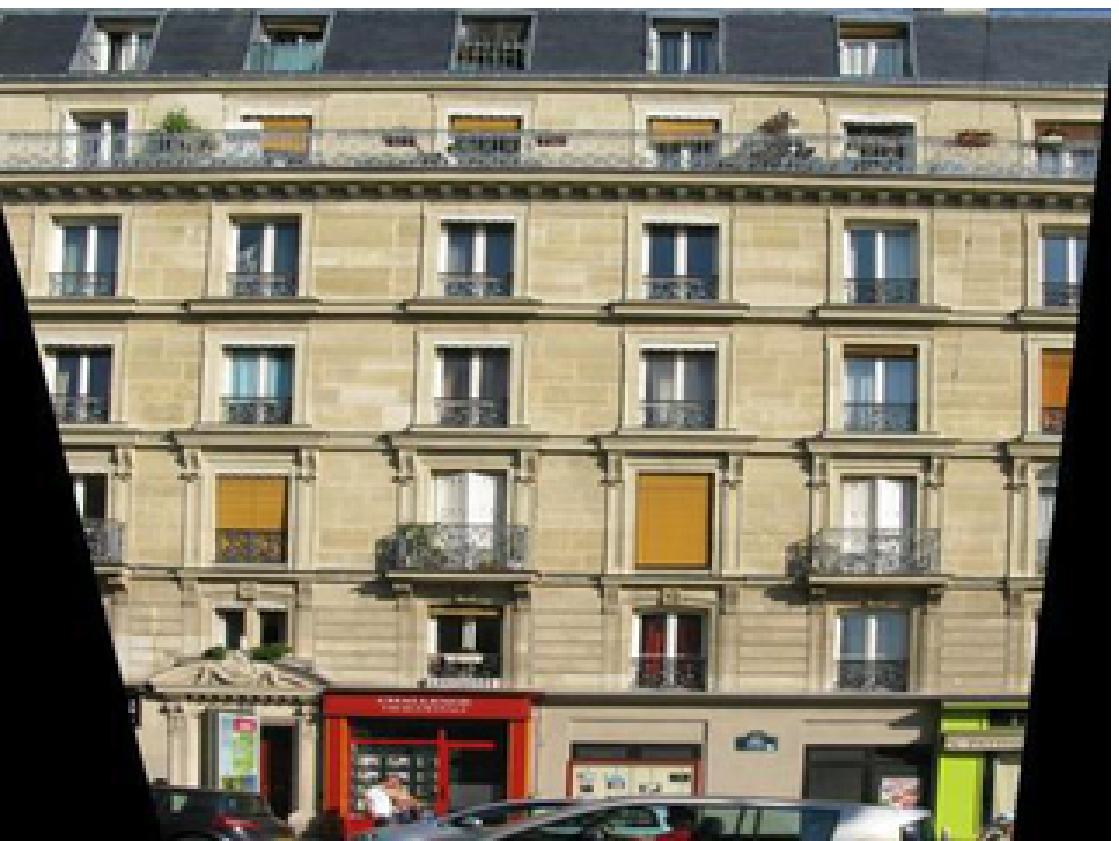}\hspace{0.02\linewidth}\includegraphics[width=0.245\linewidth,height=2.8cm]{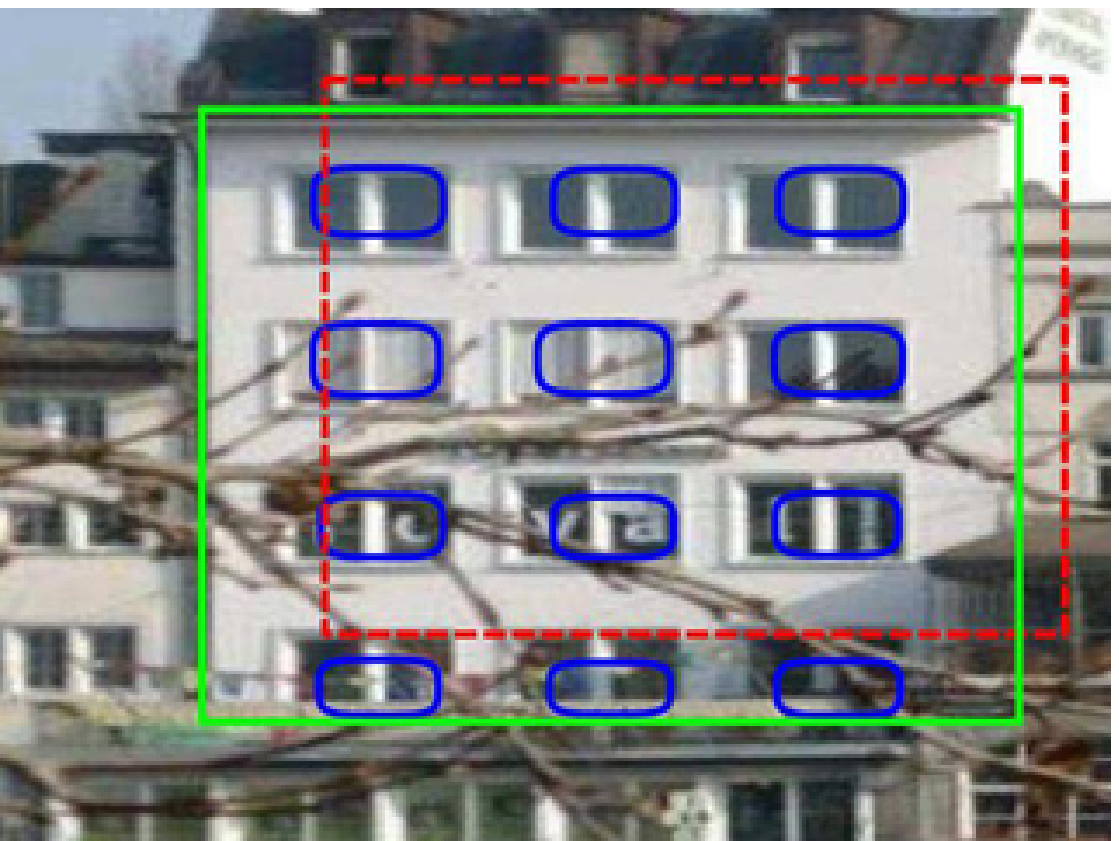}\includegraphics[width=0.245\linewidth,height=2.8cm]{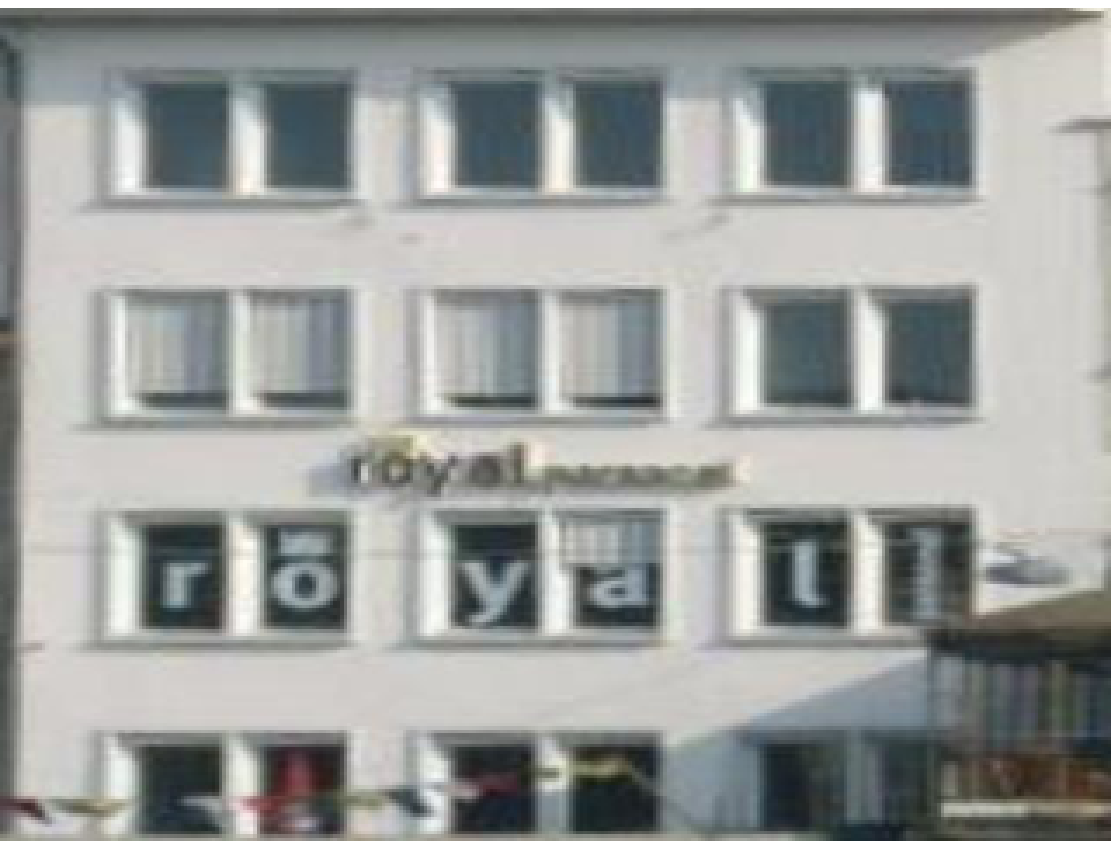}
\caption{The initial (dashed red line) and final (plain green line) registered reference boundaries overlay target images with hidden parts (left) or occlusion (right).}
\label{fig:occlusions}
\end{figure}

ZuBuD highlights other challenges as the various viewpoints cause strong changes in appearance in the rectified images especially in scale. Facades are usually poorly textured and the low resolution artifacts from the rectification make it worse. In those conditions few SIFT descriptors are extracted and they all look alike possibly causing misregistration (Fig. \ref{fig:sift}). Because the registration is bounded to the facade it can fail even if the SFM succeed relying on other contextual features. On the other hand, our approach benefits from a decent initial detection (Fig. \ref{fig:zubudres}). Occlusions are another consequence of the diversity in viewpoints. Updating the mixture weights during the EM enables our method to be robust to them as well as hidden parts (Fig. \ref{fig:occlusions}) as $\pi_{k_j}$ value can decrease if a component is not visible. Acting as a regularizer, the Dirichlet prior on mixture weights avoid complete ignorance of data by keeping the mixture weights close to their original value $\alpha_{k_j}$.

\begin{figure}[h!]
\centerline{\includegraphics[width=0.5\linewidth]{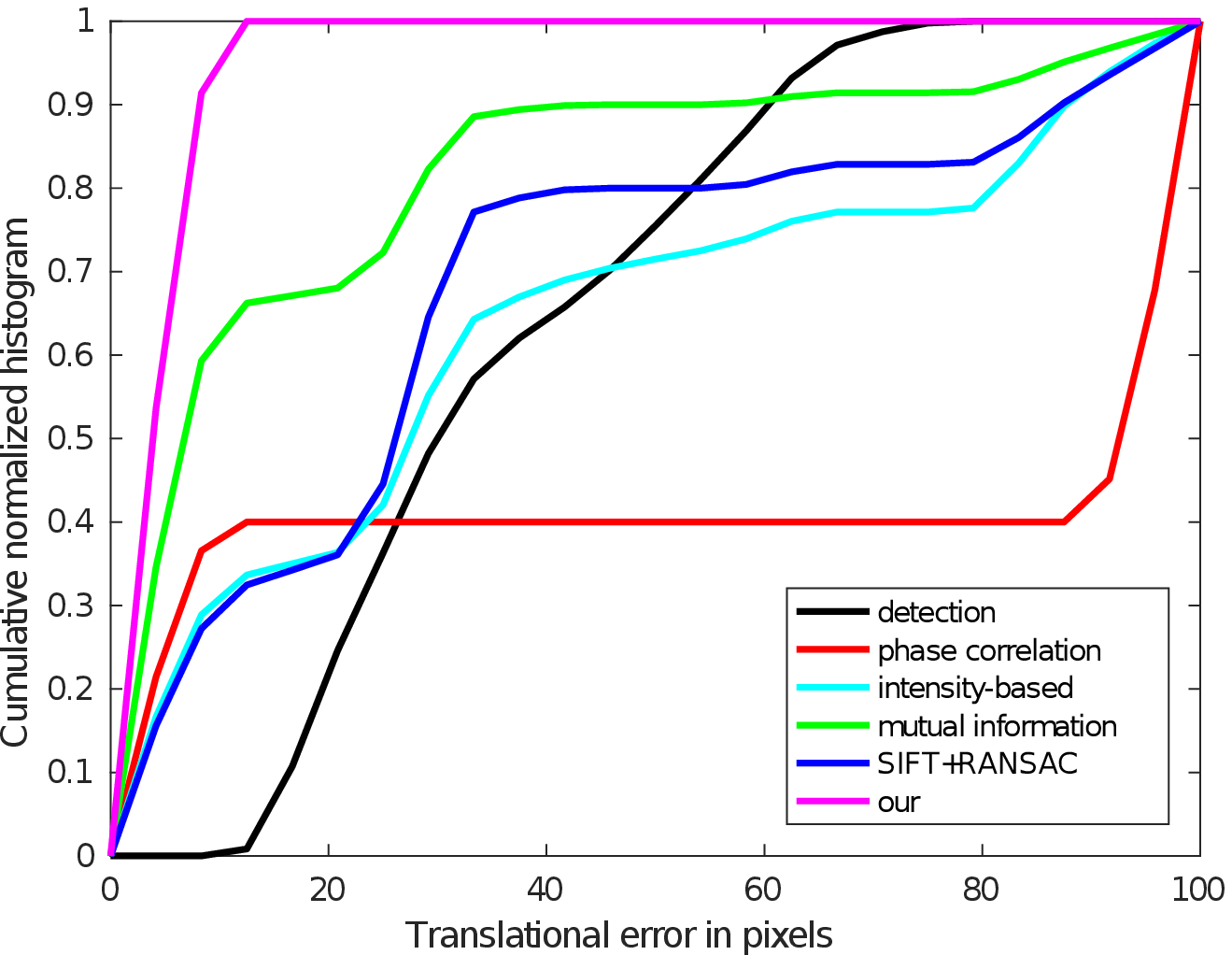}\includegraphics[width=0.48\linewidth]{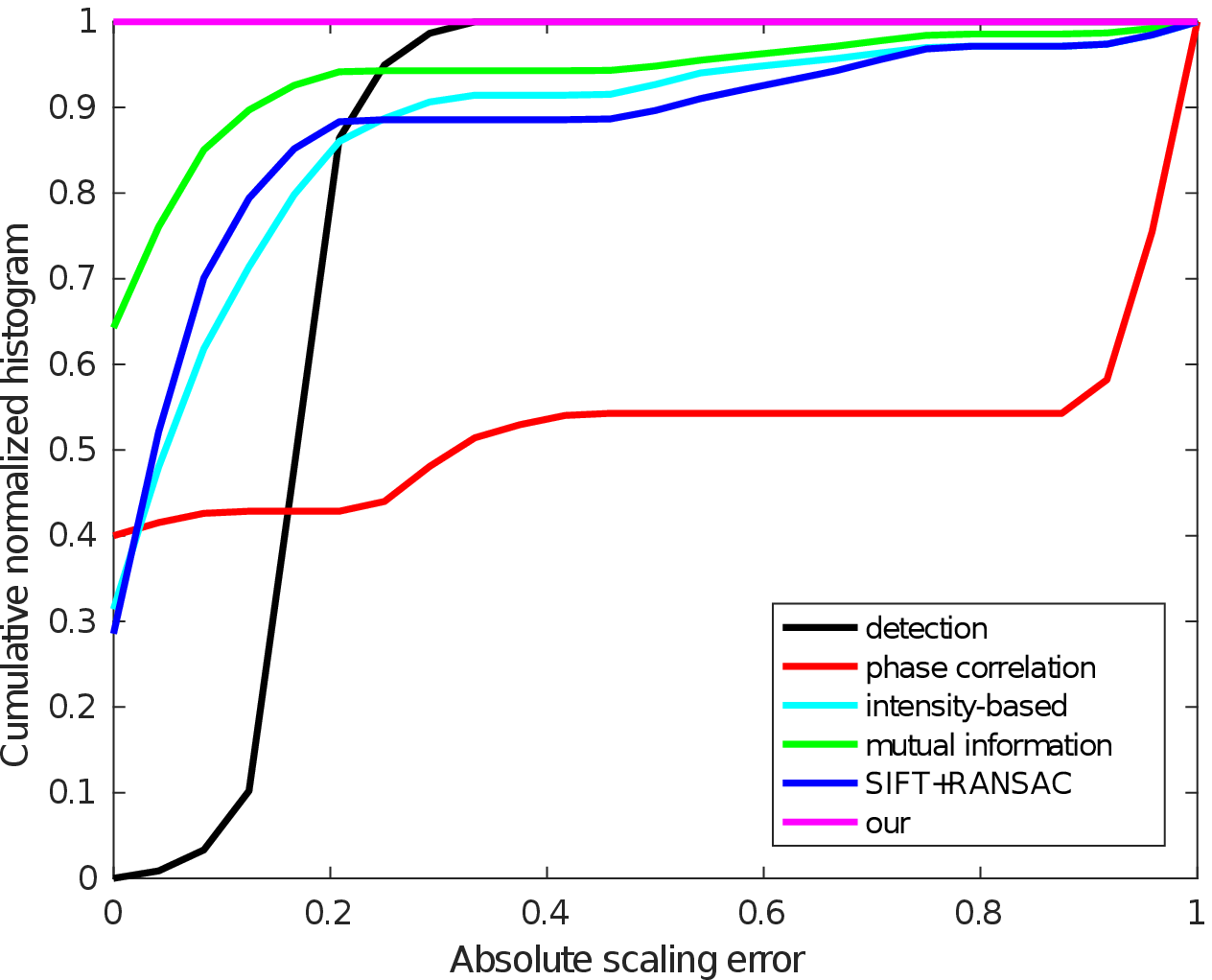}}
\caption{Registration errors in NancyLight}
\label{fig:nancyres}
\end{figure}

\begin{figure}[h!]
\centering
\includegraphics[width=0.5\linewidth,height=2.8cm]{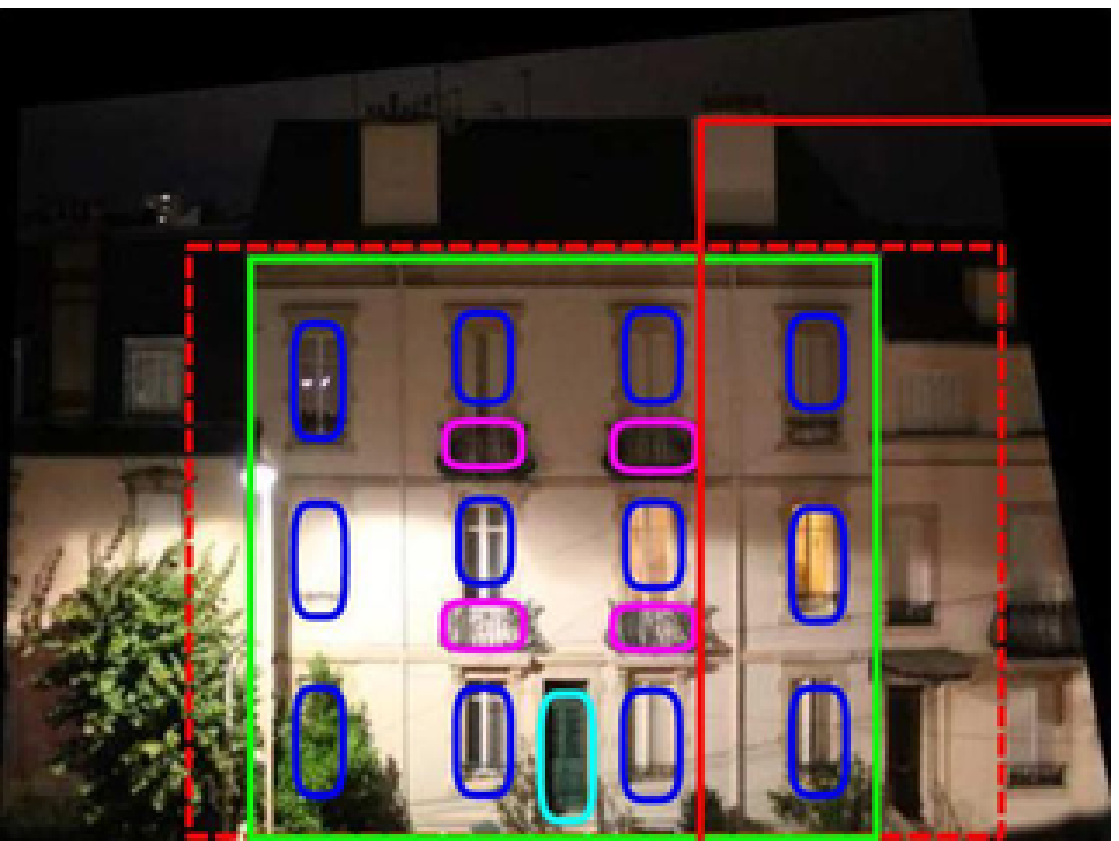}\includegraphics[width=0.5\linewidth,height=2.8cm]{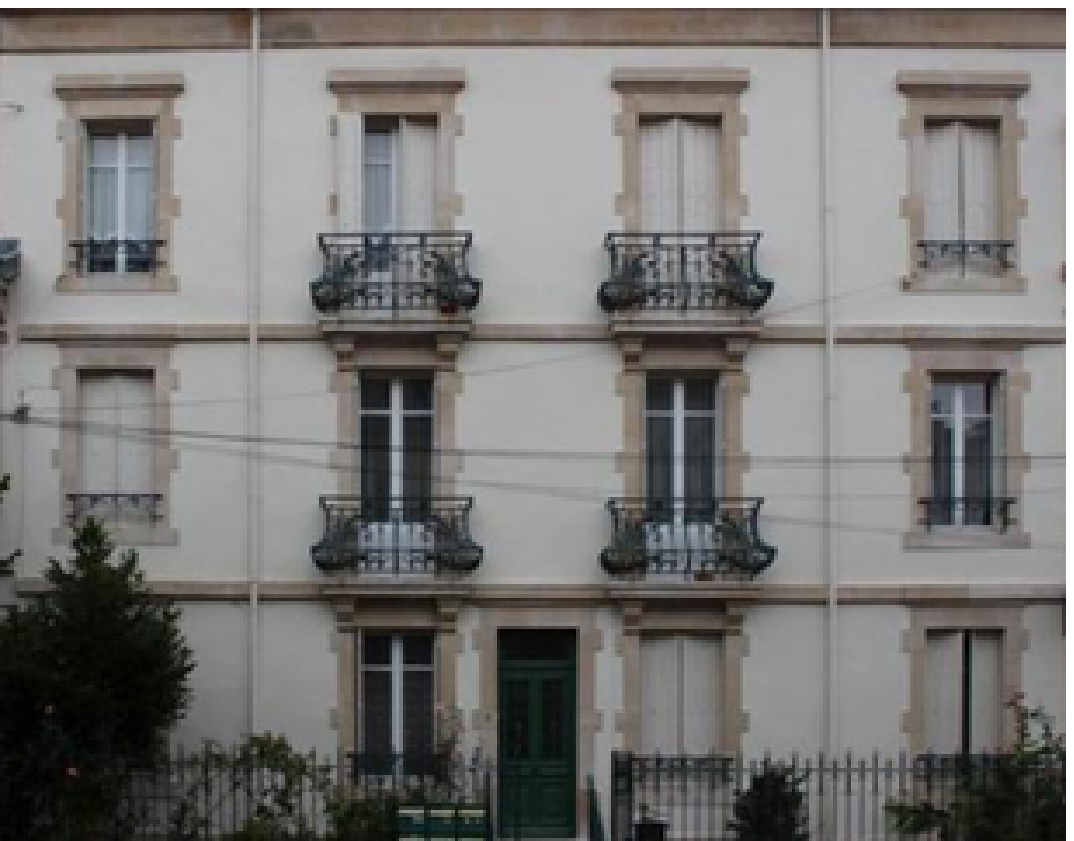}
\caption{Phase correlation approach (red) fails to estimate the registration because of the change of illumination, whereas our method (green) succeeds. The initial (dashed line) and final (plain line) registered reference boundaries overlay the target image.}
\label{fig:phcorr}
\end{figure}

The visual appearance of facades can change a lot : windows can change according to sun reflexions and to the presence of closed shutters, balconies orientation are dependent on viewpoints. If this is true on ZuBuD it is even more for the last database where the robustness to illumination changes is evaluated (Fig. \ref{fig:nancyres}). Relying on semantic segmentation enables our method to focus on the geometric structure of the facade whereas the changes in appearance are encoded in the network. The illumination invariance of the network is surprisingly good even through extreme changes in lighting that makes other methods fail (Fig. \ref{fig:phcorr}).

\begin{figure}[h!]
\centerline{\includegraphics[width=0.33\linewidth,height=2.8cm]{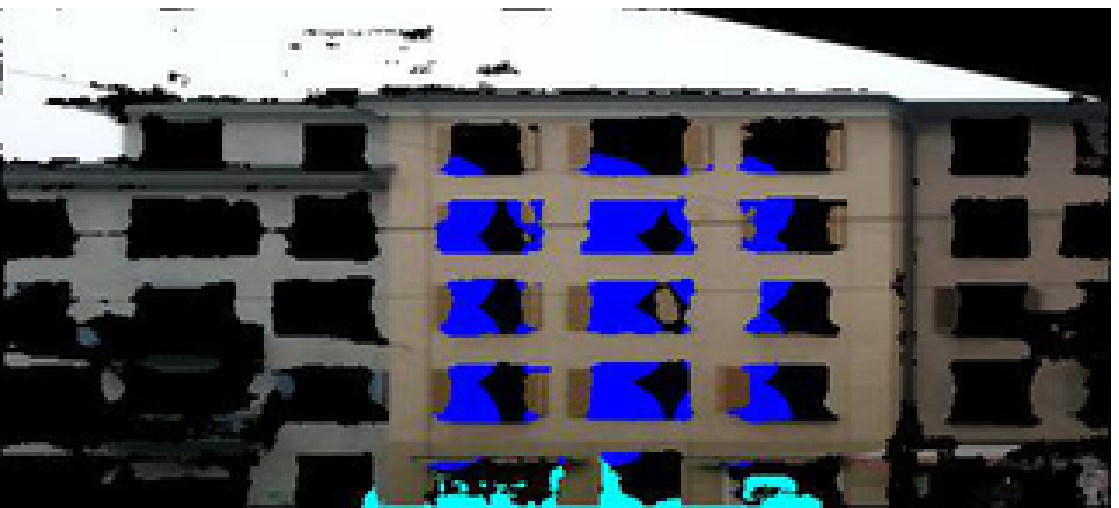}\includegraphics[width=0.33\linewidth,height=2.8cm]{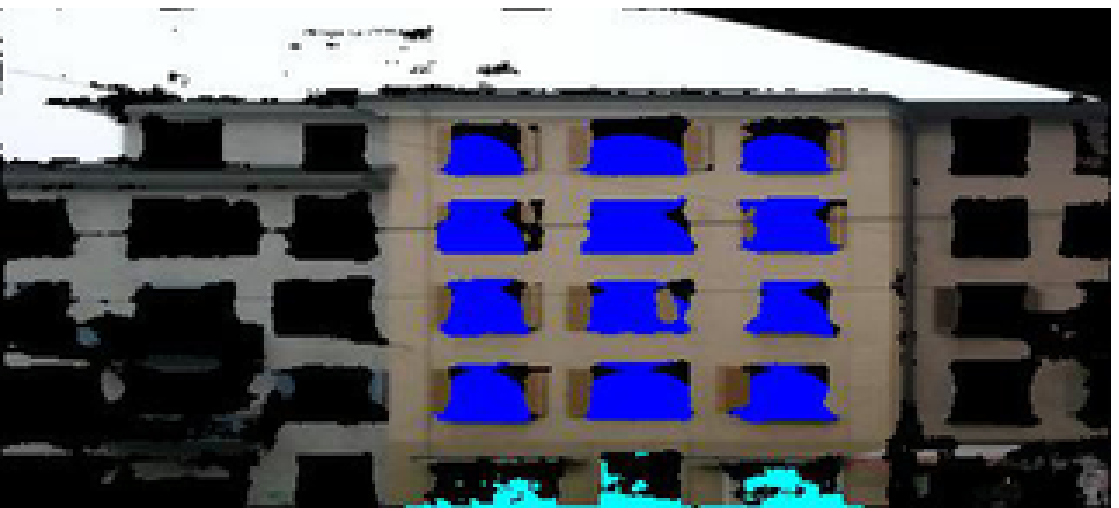}\includegraphics[width=0.33\linewidth,height=2.8cm]{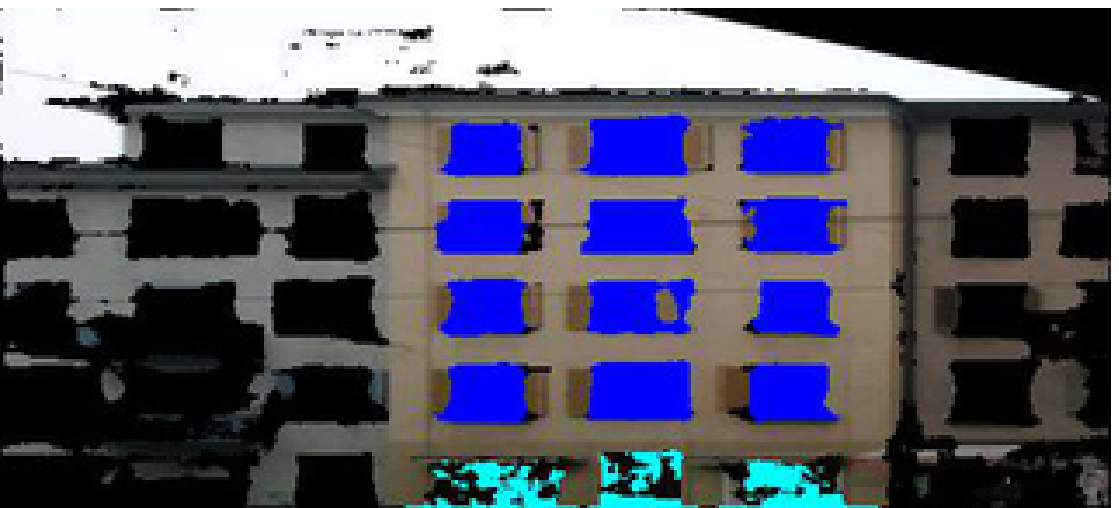}}
\caption{Evolution of the semantic segmentation during the EM on the first 3 iterations. The doors on the ground-floor are progressively correctly classified as well as they are guiding the registration.}
\label{fig:semiteration}
\end{figure}

\begin{figure}[h!]
\centerline{\includegraphics[width=0.33\linewidth,height=2.8cm]{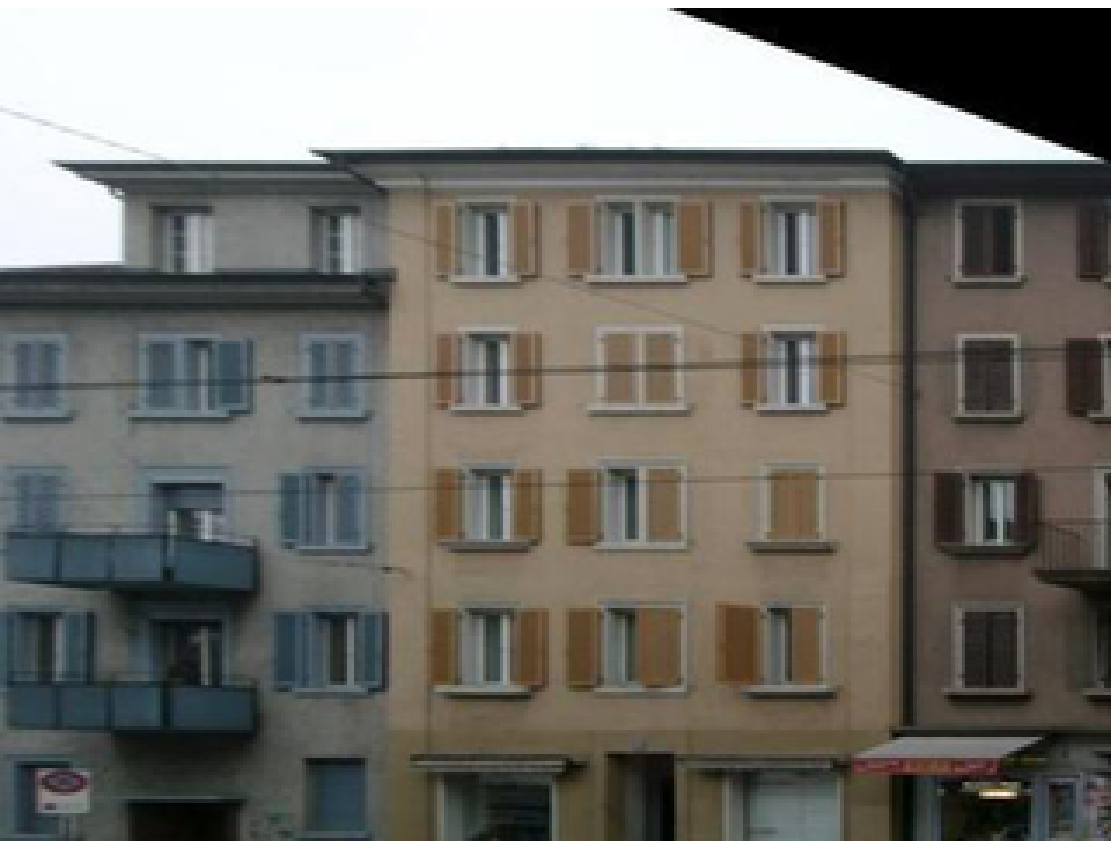}\includegraphics[width=0.33\linewidth,height=2.8cm]{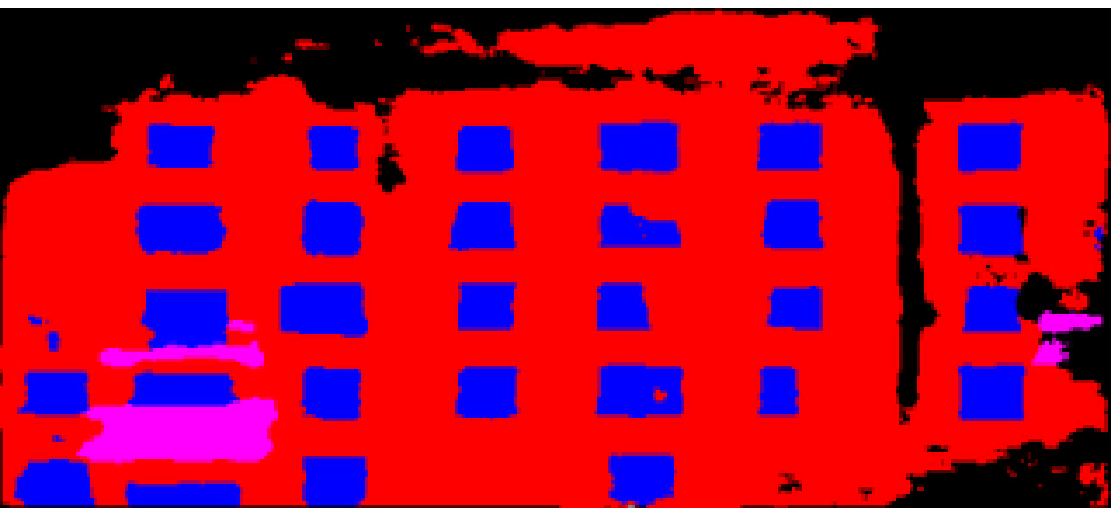}\includegraphics[width=0.33\linewidth,height=2.8cm]{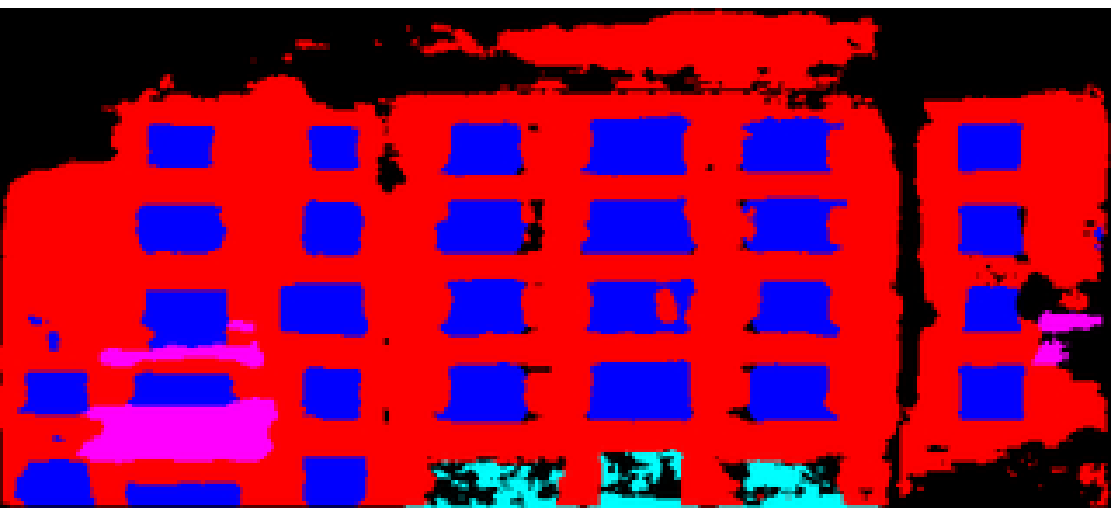}}
\caption{From left to right: the target image $I$ with the orange building as reference, the prior semantic segmentation $P(l_j|i,I)$, and the posterior semantic segmentation after registration.The 3 doors that were wrongly classified as "facade" and "window" in the prior semantic segmentation are finally correctly labeled.}
\label{fig:bettersem}
\end{figure}

\begin{table}[h!]
\tiny
\begin{tabular}{|c|c|c|c|c|c|}
\hline
& SIFT+RANSAC & phase correlation & intensity-based & mutual information & our \\
\hline
VarCity 3D & 0.04 & 0.02 & 0.37 & 0.35 & 0.03 \\
\hline
ZuBuD & 0.22 & 0.67 & 0.33 & 0.44 & 0.12 \\
\hline
\end{tabular}
\caption{Median relative errors for the 3D camera translation (relative to the distance from the building)}
\label{tab:3derror}
\end{table}
Though the semantic segmentation prior $P(l_j|i,I)$ is not updated during the EM, data points label $l_j$ can change from one iteration to another (Fig. \ref{fig:semiteration}). Indeed if misclassification is common for visually similar labels like "door" and "window", the true prior probability can be reinforced by the $L_p$ gaussian influence during registration. Eventually we can transfer the posterior probability of the data points $X$ to the prior semantic segmentation to update it (Fig. \ref{fig:bettersem}).
\subsection{Limitations}

Using ground-truth semantic references can be seen as a limitation for a real application to augmented reality or robotics. However the SegNet inference can be post-processed by introducing regularizing information based on architectural rules \cite{TeboulKSKP13}\cite{Yang2012}. These methods require the exact boundaries of the facade and can be very slow but it is perfectly suited to provide clean facade references segmented offline.

Our approach is well suited for images with sparse structures as facades but cannot be generalized to all kind of images because of spatial distributions chosen to model it ($L_p$ gaussians and uniform distribution for outliers). Cases where data points are close to a dense repartition tend to fail with data points all labeled as outliers or as a single $L_p$ gaussian if the initialization is not close enough. 


\section{Conclusion}

We have presented a bayesian model to solve jointly facade registration and semantic segmentation. The method is efficient and handle registration issues like occlusions or repetitions and improve simultaneously the semantic segmentation. As in our tests, the initialization is close enough to the solution, we can assume that the semantic segmentation inference by the CNN is stable and do not require to be reestimated online. In future work this assumption could be relaxed in the bayesian model to improve accuracy and dependence on initialization. 

  \appendix

  \section{Appendix}

  \label{app:solutions}
  With $p=2$ setting the partial derivatives of $\tilde{R}$ (cf. Eq. \ref{eq:mstep}) to zero leads to solving a polynomial system of one quadratic equation in $s$ and two linear equations in $t_x$ and $t_y$. The closed-form solution is the following :

  \begin{equation}
\begin{cases}
s = \frac{- 4 a_1  a_7 a_8 + a_3^2 a_8 + a_4^2 a_7}{2 \left( 2 a_2 a_7 a_8 - a_3 a_5 a_8 - a_4 a_6 a_7 \right)} \\
t_x = \frac{-a_3-2 a_5 s}{2 a_7} \\
t_y = \frac{-a_4-2 a_6 s}{2 a_8}
\end{cases}
\end{equation}

with
\begin{equation}
\small
\begin{split}
a_1 &= - \sum_{i,j,k_j} \beta_{i,j,k_j} \left( \frac{x_i^2}{\sigma_{k_j,x}} + \frac{y_i^2}{\sigma_{k_j,y}} \right) \\
a_2 &= \sum_{i,j,k_j} \beta_{i,j,k_j} \left( \frac{x_i \mu_{k_j,x}}{\sigma_{k_j,x}}+\frac{y_i \mu_{k_j,y}}{\sigma_{k_j,y}} \right) \\
a_3 &= 2 \sum_{i,j,k_j} \beta_{i,j,k_j} \frac{x_i}{\sigma_{k_j,x}}  \quad a_4 = 2 \sum_{i,j,k_j} \beta_{i,j,k_j}  \frac{y_i}{\sigma_{k_j,y}}  \\
a_5 &= - \sum_{i,j,k_j} \beta_{i,j,k_j}  \frac{\mu_{k_j,x}}{\sigma_{k_j,x}}  \quad a_6 = - \sum_{i,j,k_j} \beta_{i,j,k_j}  \frac{\mu_{k_j,y}}{\sigma_{k_j,y}}  \\
a_7 &= - \sum_{i,j,k_j} \beta_{i,j,k_j} / \sigma_{k_j,x} \quad a_8 = - \sum_{i,j,k_j} \beta_{i,j,k_j} / \sigma_{k_j,y}
\end{split}
\end{equation}

{\small
\bibliographystyle{ieee}
\bibliography{biblio}
}

\end{document}